\documentclass[lettersize,journal]{IEEEtran}
\usepackage{amsmath,amsfonts}
\usepackage{algorithm}
\usepackage{algpseudocode}
\usepackage{array}
\usepackage{booktabs}
\usepackage[caption=false,font=footnotesize,labelfont=rm,textfont=rm]{subfig}
\usepackage{textcomp}
\usepackage{url}
\usepackage{verbatim}
\usepackage{graphicx}
\usepackage{cite}
\usepackage{amssymb}
\usepackage{multirow}
\usepackage{float}
\usepackage{color,bm}
\usepackage{subfloat}
\usepackage{subfig}

\newtheorem{remark}{Remark}
\begin{document}

\title{cc-DRL: a Convex Combined Deep Reinforcement Learning Flight Control Design for a Morphing Quadrotor}
\author{Tao Yang, Huai-Ning Wu, and Jun-Wei Wang
        % <-this % stops a space
%\thanks{This paper was produced by the }% <-this % stops a space
\thanks{Tao Yang is with the School of Automation Science and Electrical Engineering, Beihang University, Beijing 100191, China (e-mail: tiantaoyang@buaa.edu.cn).}
\thanks{Huai-Ning~Wu is with the Science and Technology on Aircraft Control Laboratory, School of Automation Science and Electrical Engineering, Beihang University, Beijing 100191, China, and also with Hangzhou International Innovation Institute of Beihang University, Hangzhou 311115, China (e-mail: whn@buaa.edu.cn).}
\thanks{Jun-Wei~Wang is with the School of Intelligence Science and Technology, University of Science and Technology Beijing, Beijing 100083, China (e-mail: junweiwang@ustb.edu.cn).}
}

% The paper headers
% \markboth{Journal of \LaTeX\ Class Files,~Vol.~14, No.~8, August~2021}%
% {Shell \MakeLowercase{\textit{et al.}}: A Sample Article Using IEEEtran.cls for IEEE Journals}

% \IEEEpubid{0000--0000/00\$00.00~\copyright~2021 IEEE}
% \IEEEpubidadjcol
% Remember, if you use this you must call \IEEEpubidadjcol in the second
% column for its text to clear the IEEEpubid mark.

\maketitle

\begin{abstract}
In comparison to common quadrotors, the shape change of morphing quadrotors endows it with a more better flight performance but also results in more complex flight dynamics. Generally, it is extremely difficult or even impossible for morphing quadrotors to establish an accurate mathematical model describing their complex flight dynamics. To figure out the issue of flight control design for morphing quadrotors, this paper resorts to a combination of model-free control techniques (e.g., deep reinforcement learning, DRL) and convex combination (CC) technique, and proposes a convex-combined-DRL (cc-DRL) flight control algorithm for position and attitude of a class of morphing quadrotors, where the shape change is realized by the length variation of four arm rods. In the proposed cc-DRL flight control algorithm, proximal policy optimization algorithm that is a model-free DRL algorithm is utilized to off-line train the corresponding optimal flight control laws for some selected representative arm length modes and hereby a cc-DRL flight control scheme is constructed by the convex combination technique. Finally, simulation results are presented to show the effectiveness and merit of the proposed flight control algorithm.
\end{abstract}


\section{Introduction}
\label{sec:Introduction}
\IEEEPARstart{A}{s} a class of well-mature platforms, quadrotor unmanned aerial vehicles (UAVs) provide mobilities in cluttered or dangerous environments where the human being is at risk and are helpful for many civilian and military applications such as surveillance of forest fire detection, high building inspection, battlefield monitor, and battlefield weapon delivery, etc. Over the past few decades, the robotics community has experienced a very active and prolific topic in quadrotors and breakthroughs have been made for the issues of control algorithms, architectural design and applications \cite{idrissi2022review, amin2016review, lopezsanchez2023pid}. In above issues, flight control algorithms implicitly determine the performance of the quadrotors. Hence, the issue of flight control scheme design for quadrotors is very significant. This issue is extremely difficult since a fact that quadrotors present highly nonlinear and coupled dynamics that can be stabilized using four control inputs. This fact has also promoted the attention of many control practitioners and theoretical specifics \cite{amin2016review, lopezsanchez2023pid, 9552618}.

After years of developments, common quadrotors have been commercialized and their technologies have become more and more mature. Yet quadrotors must sometime fly through narrow gaps in disaster scenes in geographical investigations and even on battlefields. Hence it is very useful for quadrotors that can change their shapes. At the same time, the shape change endows quadrotors with stronger environmental adaptability and more complex task completion \cite{hu2021design}. Three types of morphing quadrotors have been reported in the existing works: tiltrotor quadrotor, multimodal quadrotor, and foldable quadrotor \cite{patnaik2021towards}. For the tiltrotor quadrotor \cite{al2020state}, the input dimension of the control forces is extended to enhance its maneuverability by changing the direction of the rotor axis. The rotor lift force direction is thereby changed for quadrotors and additional design of the tilt controller is thus required. Both a MIMO PID flight controller \cite{dos2023cascade} and an ADRC (active disturbance rejection control) flight controller \cite{shen2023adaptive} are reported for a tiltrotor quadrotor with a better robustness performance. For the multimodal quadrotor \cite{tan2021survey}, the quadrotor can perform different tasks by presetting several variation modes, and switching among them during flight to meet the multitasking requirements. To this end, for each variation mode, a corresponding control law is predesigned \cite{gao2023multimode, rao2023puffin}. For the foldable quadrotor \cite{yang2019design}, the quadrotor modifies its size by actively changing the mechanical structure to enhance its passability (e.g., passing narrow channels). To ensure the flight safety of the foldable quadrotor, the change of mechanical structure is considered as a model perturbation and then a robust control law is designed \cite{patnaik2023adaptive, jia2023aerial, wu2023ring}. Despite the above progresses, the aforementioned flight control algorithms are developed by the matured model-based control theory and thus lack of learning ability. %The shape change endows the morphing quadrotors with high maneuverability, but it also comes with more complex flight dynamics. Hence, it is difficult or even impossible to establish a mathematical model describing the complex flight dynamics of morphing quadrotors.

With the rapid development of artificial intelligence (AI), deep reinforcement learning (DRL) combines the representation ability of deep learning (DL) and the decision ability of reinforcement learning (RL) \cite{arulkumaran2017deep}, \cite{luong2019applications}, which has a strong exploratory ability to solve complex dynamic planning problems, and its performance in solving optimal control problems is becoming more and more significant \cite{kiran2021deep}. In the last ten years, RL/DRL has been successfully used to solve the optimal control problem of quadrotor dynamics \cite{cheng2023optimal, song2023reaching, hwangbo2017control, lopes2018intelligent, koch2019reinforcement, bernini2021few, jiang2021quadrotor, bernini2024reinforcement}, where the strong learning and exploration ability of DRL solves the challenges posed by the strong nonlinearity in quadrotor dynamics. In \cite{cheng2023optimal, song2023reaching, 9789160}, RL-based approximate optimal flight control schemes were proposed for position and attitude of a quadrotor. DRL-based approximated optimal flight control laws were proposed for position and attitude of quadrotors \cite{hwangbo2017control, lopes2018intelligent, koch2019reinforcement, bernini2021few, jiang2021quadrotor, bernini2024reinforcement}. Note that the aforementioned results only focus on flight control design of common quadrotors. To the best of authors' knowledge, the research on DRL-based flight control design of for morphing quadrotors is quite few. % and may be an open question.

\begin{figure}[!ht]
  \centering
  \includegraphics[width=1\columnwidth]{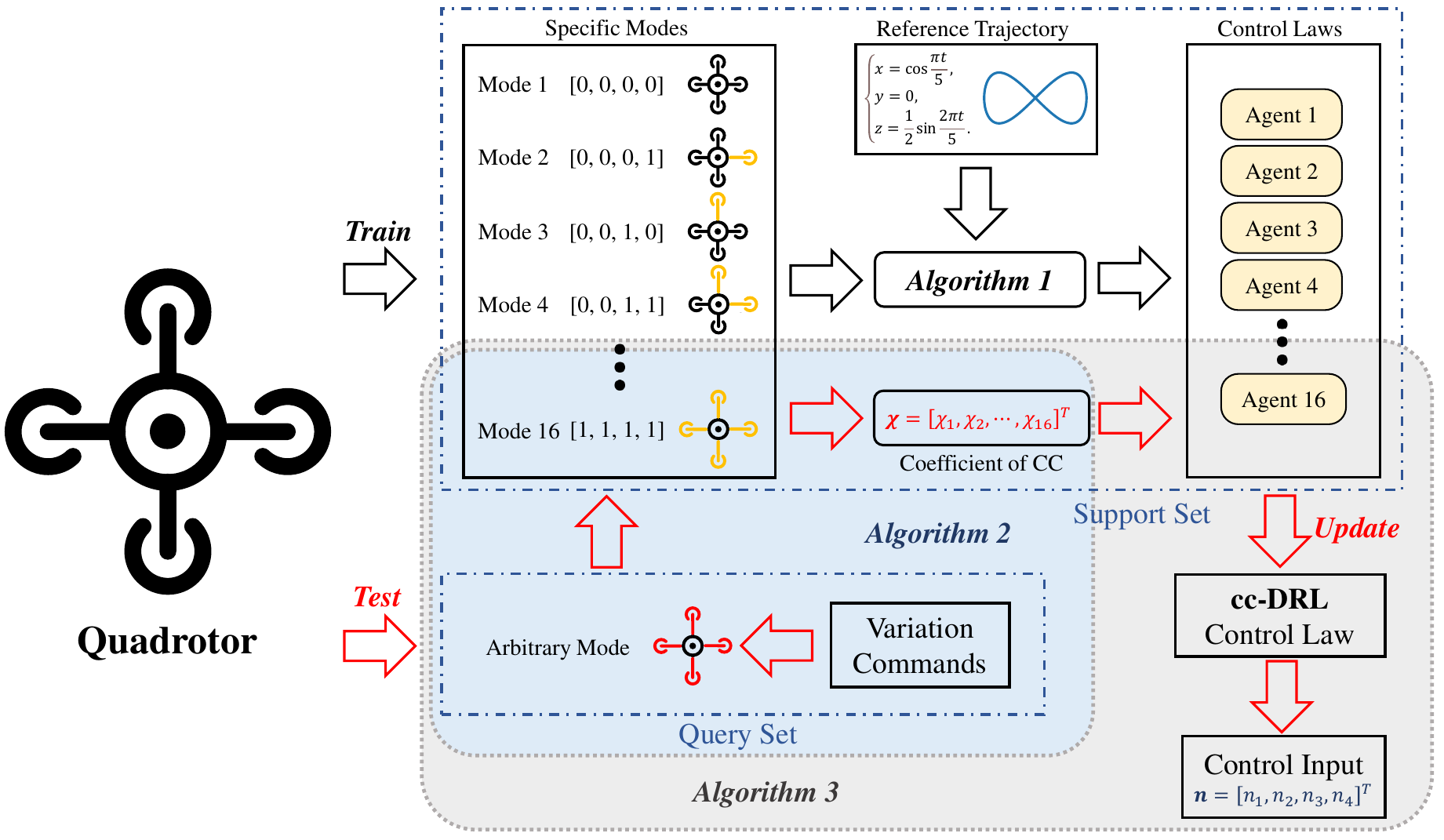}
  \caption{
    The structure of the proposed cc-DRL flight control algorithm for an arm-rod-length-varying quadrotor.
     \textbf{Algorithm 1} shows the elaborate DRL algorithm for off-line training the optimal flight control laws for some selected representative length modes of four arm rods.
    \textbf{Algorithm 2} proposes a convex combination method for arbitrary length of four arm rods, which can be used online or substituted by an offline pretrained neural network.
     \textbf{Algorithm 3} provides a cc-DRL flight control scheme that receives external length variation commands (query set) for four arm rods and online updates the combination weight values of the trained optimal flight control laws (support set) to achieve a near optimal flight performance.
     }
  \label{map}
\end{figure}

In this study, the issue of optimal flight control design is addressed for position and attitude of a class of morphing quadrotors, where the shape change is carried out via the length variation of four arms. With the aid of a combination of DRL and convex combination (CC), a convex-combined-DRL (cc-DRL) flight control algorithm is proposed by taking full account of the transition process in length variation for four arms to endow the morphing quadrotor with a better flight performance. In the proposed cc-DRL flight control algorithm, some representative arm length modes are first chosen for length variation of four arm rods. For each specific arm length mode of four arm rods, a corresponding optimal flight control scheme is then off-line trained by a proximal policy optimization (PPO) algorithm that is a model-free DRL algorithm. By interpolation of these off-line trained optimal flight control laws in the CC framework, an online overall flight control scheme is proposed and thus named as a cc-DRL one, where the ideal combination weight values are the solution to the non-convex quadratic programming problem that is iteratively solved by the sequential least square programming algorithm. Fig. \ref{map} shows the structure of the proposed cc-DRL flight control algorithm.

The main contribution and key novelty of this study lie in that a cc-DRL flight control scheme for position and attitude of an arm-rod-length-varying quadrotor assisted by a combination of DRL and CC technique. Essentially, the proposed cc-DRL flight control algorithm is a model-free one due to the introduction of PPO algorithm. That is to say, different from the existing works \cite{dos2023cascade, shen2023adaptive, tan2021survey, gao2023multimode, rao2023puffin, yang2019design, patnaik2023adaptive, jia2023aerial, wu2023ring}, this study develops a pure data-driven flight control algorithm for the arm-rod-length-varying quadrotor without any model knowledge of flight dynamics. On the other hand, the morphing quadrotor addressed in this study is completely different from the common one discussed in \cite{hwangbo2017control, lopes2018intelligent, koch2019reinforcement, bernini2021few, jiang2021quadrotor, bernini2024reinforcement}. Furthermore, the shape change of the morphing quadrotor introduces more complex flight dynamics in comparison to the common one.

The remainder of this paper is organized as follows. Section~\ref{sec:Background} introduces some background of morphing quadrotor dynamics, control objective, and PPO algorithm. In Section~\ref{sec:Deep Reinforcement Learning for Control Law Designing}, a PPO-based off-line optimal flight control design is introduced for some selected representative arm length modes. Then, a cc-DRL flight control scheme is presented in Section~\ref{sec:Online Adaptive Control Law} by the off-line trained optimal flight control laws and the CC technique. Performance evaluation results are presented in Section~\ref{sec:Simulation} to support the proposed cc-DRL flight control algorithm, and conclusions follow in Section~\ref{sec:Conclusion}.

\section{Preliminaries and Problem Formulation}
\label{sec:Background}
\subsection{Morphing quadrotor dynamics}
\label{sec:Quadrotor Dynamics}
\begin{figure}[!ht]
\centering
\includegraphics[width=0.8\columnwidth]{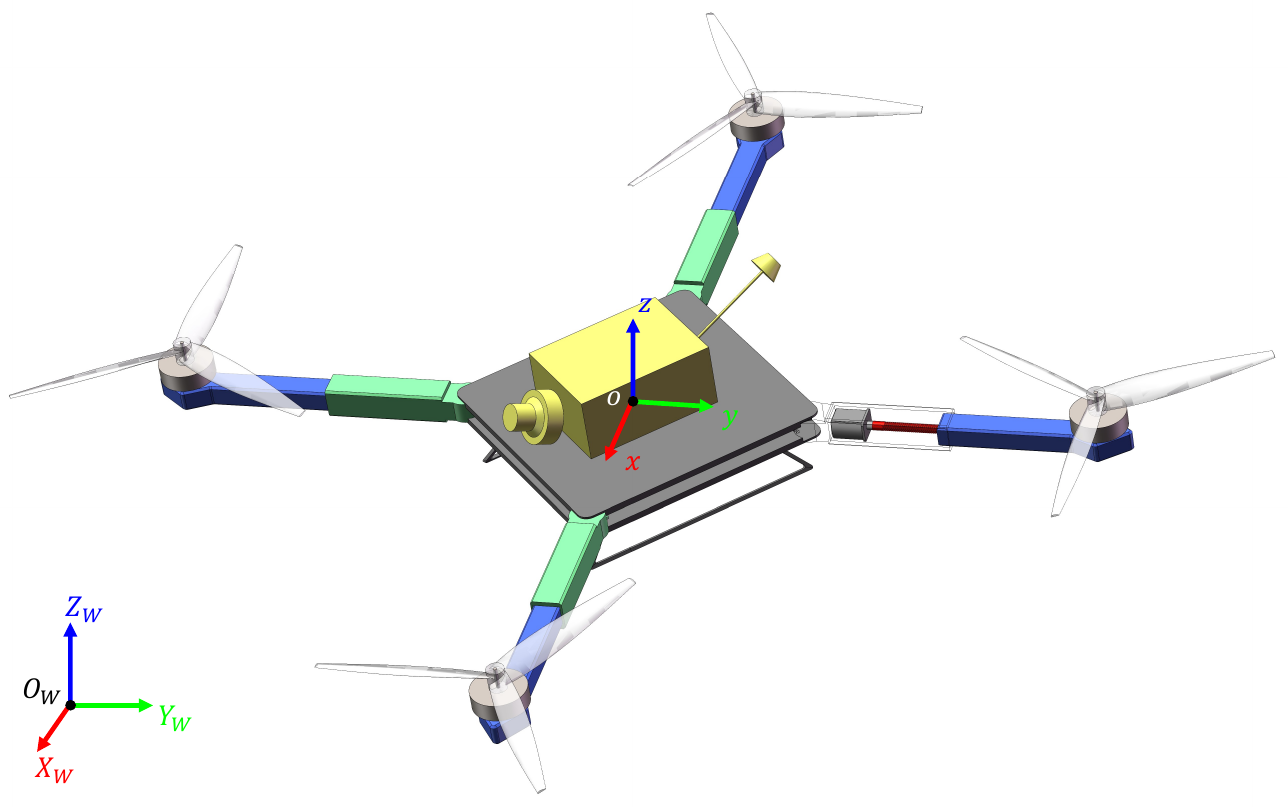}
\caption{Sketch map of a morphing quadrotor with four variable-length arm rods.}
\label{fig1}
\end{figure}

A morphing quadrotor addressed in this paper has four variable-length arm rods and its sketch map is shown in Fig. \ref{fig1}. In the addressed morphing quadrotor, each arm rod can independently change its length in response to the change of flight environment and missions. Hence, four variable-length arm rods endow the morphing quadrotor with a better adaptability of flight environments and unplanned multipoint missions. But the independent length change of four arm rods changes the mass distribution of the morphing quadrotor and disrupts the symmetric structure of the conventional quadrotor. Flight dynamics of the morphing quadrotor are more complex than the one of the common quadrotor. Essentially, morphing quadrotors are a class of reconfigurable systems. % On the other hand,

% a mathematical model will be established on the basis of the symmetric quadrotor \cite{idrissi2021modelling}, \cite{hossain2016multi} for the sake of numerical simulation, where length variations of four arm rods (i.e., $l_i(t)$, $i\in\{1,2,3,4\}$) are considered as time-varying parameters. To do this,
To capture such complex flight dynamics, two frames are introduced: a world internal frame $F_W: \{O_W,X_W,Y_W,Z_W\}$ and a moving frame $F_B: \{o,x,y,z\}$ attached to the quadrotor body at its mass center (see Fig. \ref{fig1}). The rotational matrix between the moving frame $F_B$ and the world internal one $F_W$ is chosen as follows
\begin{equation}\label{eq:1}
  R_{B}^{W} =
  \begin{bmatrix}
    c_{\theta}c_{\psi} & s_{\phi}s_{\theta}c_{\psi}-c_{\phi}s_{\psi} & c_{\phi}s_{\theta}c_{\psi}+s_{\phi}s_{\psi} \\
    c_{\theta}s_{\psi} & s_{\phi}s_{\theta}s_{\psi}+c_{\phi}c_{\psi} & c_{\phi}s_{\theta}s_{\psi}-s_{\phi}c_{\psi} \\
    -s_{\theta} & s_{\phi}c_{\theta} & c_{\phi}c_{\theta}
  \end{bmatrix}
\end{equation}
where $s_{(\cdot)}=\sin(\cdot)$ and $c_{(\cdot)}=\cos(\cdot)$ are the respective sine and cosine, and $\phi$, $\theta$, and $\psi$ are the quadrotor's attitude angles.

In the morphing quadrotor, four rotors are respectively fixed at the end of four arm rods. Angular velocities of these four rotors are denoted by $n_{i}$, $i\in\{1,2,3,4\}$ and chosen as manipulated control inputs, i.e., $\textit{\textbf{u}}\triangleq[n_1\ n_2\ n_3\ n_4]^T$. Both mass center position vector $\textit{\textbf{x}}\triangleq[x\ y\ z]^T\in\mathbb{R}^3$ and attitude angle vector $\bm{\varpi}\triangleq[\phi\ \theta\ \psi]^T\in\mathbb{R}^3$ are chosen as state variables of the morphing quadrotor. The evolution dynamics of these state variables is governed by the following nonlinear system model
\begin{equation}\label{eq:2}
  \begin{bmatrix}
    \ddot{x} \\ \ddot{y} \\ \ddot{z} \\ \ddot{\phi} \\ \ddot{\theta} \\ \ddot{\psi}
  \end{bmatrix}
  =
  \begin{bmatrix}\vspace{1ex}
    f_1(\cdot) \\\vspace{1ex}
    f_2(\cdot) \\\vspace{1ex}
    f_3(\cdot) \\\vspace{1ex}
    f_4(\cdot) \\\vspace{1ex}
    f_5(\cdot) \\
    f_6(\cdot)
  \end{bmatrix}
\end{equation}
where $f_i(\cdot)$, $i\in\{1,2,\cdots,6\}$ are functions of the parameters $\textit{\textbf{x}}$, $\dot{\textit{\textbf{x}}}$, $\bm{\varpi}$, $\dot{\bm{\varpi}}$, $\textit{\textbf{u}}$, $m$, $I_x(t)$, $I_y(t)$, $I_z(t)$, $l_1(t)$, $l_2(t)$, $l_3(t)$, and $l_4(t)$, in which $m$ is the quadrotor mass, {$I_{x}(t)$, $I_{y}(t)$, and $I_{z}(t)$} are inertia moments of the quadrotor, and the time-varying parameters $l_j(t)$, $j\in\{1,2,3,4\}$ are used to describe the dynamic changes in the length of four arm rods.

%i.e.,  $$f_i(\textit{\textbf{x}},\dot{\textit{\textbf{x}}},\bm{\varpi},\dot{\bm{\varpi}},\textit{\textbf{u}},m, {I_x(t),I_y(t),I_z(t)},l_j(t),l_2(t),l_3(t),l_4(t)),$$
\subsection{Control objective}
\label{sec:Goal of Control Law Designing}
Let $\textit{\textbf{x}}_r$ be a preset flight path of the morphing quadrotor. The corresponding position tracking error vector $\tilde{\textit{\textbf{x}}}$ is defined by $\tilde{\textit{\textbf{x}}}\triangleq {\textit{\textbf{x}}}-{\textit{\textbf{x}}}_r$. To fully describe the quadrotor's dynamics, a new 12-dimensional state vector $\textit{\textbf{s}}$ is introduced and defined as
\begin{equation}\label{eq:6}
  \boldsymbol{s}\triangleq
  \begin{bmatrix}
    \tilde{\textit{\textbf{x}}}^T\quad \bm{\varpi}^T\quad \dot{\tilde{\textit{\textbf{x}}}}^T\quad \dot{\bm{\varpi}}^T
  \end{bmatrix}^T
  \in \boldsymbol{\mathcal{S}}
\end{equation}
where $\boldsymbol{\mathcal{S}}$ is the state space, i.e., the set of all possible 12-dimensional {state} vectors of the quadrotor. These $12$ {states} include the position tracking error vector $\tilde{\textit{\textbf{x}}}$, the attitude angle vector $\bm{\varpi}$, the linear velocity error vector $\dot{\tilde{\textit{\textbf{x}}}}$, and the attitude angular velocity vector $\dot{\bm{\varpi}}$.

The control objective of this paper is to find an approximate solution to the optimal flight control problem (\ref{eq:4}) for the morphing quadrotor such that the quadrotor flies along the preset flight path $\boldsymbol{x}_{r}$ with a minimal energy consumption.
\begin{equation}\label{eq:4}
 \boldsymbol{u^{*}}(\boldsymbol{s})=\underset{\boldsymbol{u}}{\text{argmin}}\ J
\end{equation}
where $\boldsymbol{u^{*}}(\boldsymbol{s})$ is the optimal flight control law and $J$ is the performance metric of the above optimal flight control problem and defined by
\begin{equation}\label{eq:5}
  J=\Phi\left[\boldsymbol{s}\left(t_f\right), t_f\right]+\int_{t_{0}}^{t_{f}}F\left[\boldsymbol{s}\left(t\right), \boldsymbol{u}\left(t\right), t\right]\, \text{d}t
\end{equation}
in which $t_{0}$ is the initial time, $t_{f}$ is the terminal time, $\int_{t_{0}}^{t_{f}}F\left[\boldsymbol{s}\left(t\right), \boldsymbol{u}\left(t\right), t\right]\, \text{d}t$ is an integral performance metric, and $\Phi\left[\boldsymbol{s}\left(t_f\right), t_f\right]$ is a terminal performance metric, respectively. A detailed design process of the performance metric (\ref{eq:5}) will be discussed in Section \ref{sec:Reward Function Design}.

Due to a fact that physical mechanisms of the morphing quadrotor with four variable-length arm rods are still unclear and lack domain knowledge, it is difficult or even impossible to obtain an accurate mathematical model of the form \eqref{eq:2}. The existing mature model-based RL algorithms are unable to solve the optimal flight control problem (\ref{eq:4}). In this situation, this paper will resort to a DRL algorithm, which is a type of model-free RL algorithm. The DRL algorithm will be used to train a policy function and get a nonlinear state-feedback optimal controller from the real-time flight state data \cite{chow2019lyapunov}. The obtained optimal flight controller guides the morphing quadrotor to fly along the preset path with a better performance. Note that both state space and action space of the optimal quadrotor flight control are continuous, PPO algorithm will be utilized to train the DRL-based optimal flight control scheme.

\subsection{Proximal policy optimization (PPO) algorithm}
\label{sec:Deep Reinforcement Learning}
PPO algorithm is a model-free DRL algorithm \cite{schulman2017proximal}. A state value function is introduced to describe the {value of state} $\boldsymbol{s}$, which is computed as follows
\begin{equation}\label{eq:14}
  \begin{split}
    V^{\pi}\left(\boldsymbol{s}\right)
    &=\mathbb{E}_{\pi}\left[G_{t}|\boldsymbol{s}_{t}=\boldsymbol{s}\right]\\
    &=\mathbb{E}_{\pi}\left[r_{t+1}+\gamma V^{\pi}\left(\boldsymbol{s}_{t+1}\right)|\boldsymbol{s}_{t}=\boldsymbol{s}\right]
  \end{split}
\end{equation}
where $G_{t} \in \mathbb{R}$ is the accumulated rewards of a trajectory generated from state $\boldsymbol{s}_{t}=\boldsymbol{s}$ guided by the policy $\pi$,
 $\gamma \in \left(0, 1\right)$ is the discount factor of the reward, $r_{t+1} \in \mathbb{R}$ is the reward of the next state $\boldsymbol{s}_{t+1}$,
 and $\mathbb{E}_{\pi}$ represents the expectation of policy $\pi$. The goal of DRL is to find a policy function such that the sequential decisions of the agent have the maximum accumulated rewards, i.e., maximum of the expectation of the initial state value function $J_{\boldsymbol{s}_{1}}$ by choosing an appropriate policy $\pi$:
\begin{equation}\label{eq:15}
  J_{\boldsymbol{s}_{1}}
  =\mathbb{E}_{\boldsymbol{s}_{1}\sim p\left(\boldsymbol{s}_{1}\right)}\left[V^{\pi}\left(\boldsymbol{s}\right)\right]
\end{equation}
where $\boldsymbol{s}_{1} \in \mathbb{R}^{12}$ is the initial state, $p:\mathbb{R}^{12} \mapsto \mathbb{R}$ is the distribution function of initial state in state space $\boldsymbol{\mathcal{S}}$, and $V^{\pi}\left(\boldsymbol{s}_{1}\right)$ is the state value function of $\boldsymbol{s}_{1}$ guided by the policy $\pi$.

An action value function of state-action pair is adopted to describe the value of a policy $\boldsymbol{a}$, i.e., the value of action $\boldsymbol{a}$ at state $\boldsymbol{s}$, which can be computed as follows:
\begin{equation}\label{eq:16}
  Q^{\pi}\left(\boldsymbol{s}, \boldsymbol{a}\right)
    =\mathbb{E}_{\pi}\left[G_{t}|\boldsymbol{s}_{t}=\boldsymbol{s}, \boldsymbol{a}_{t}=\boldsymbol{a}\right]
\end{equation}
The relationship between the state value function and the action value function is represented as follows
\begin{equation}\label{eq:17}
  \begin{split}
    V^{\pi}\left(\boldsymbol{s}\right)
    &=\mathbb{E}_{\boldsymbol{a} \sim \pi \left(\boldsymbol{a}|\boldsymbol{s}\right)}
    \left[Q^{\pi}\left(\boldsymbol{s}, \boldsymbol{a}\right)\right]\\
    &=\sum_{\boldsymbol{a} \in \boldsymbol{\mathcal{A}}}
    \pi \left(\boldsymbol{a}|\boldsymbol{s}\right)Q^{\pi}\left(\boldsymbol{s}, \boldsymbol{a}\right)
  \end{split}
\end{equation}
where $\pi \left(\boldsymbol{a}|\boldsymbol{s}\right)$ represents the probability distribution of the action $\boldsymbol{a}$ at state $\boldsymbol{s}$ guided by the policy $\pi$. To facilitate the policy optimization, advantage function of action is introduced and is calculated as follows
\begin{equation}\label{eq:18}
  A^{\pi}\left(\boldsymbol{s}, \boldsymbol{a}\right)
  =Q^{\pi}\left(\boldsymbol{s}, \boldsymbol{a}\right)-V^{\pi}\left(\boldsymbol{s}\right)
\end{equation}
which describes the advantage of action $\boldsymbol{a}$ at state $\boldsymbol{s}$ over the average based on policy $\pi$.

For the agent with a better decision, it is desired that the action with a larger advantage has a higher probability to be selected and the one with a smaller advantage has a lower probability to be selected. Following this idea to optimize the policy function, the optimization goal that needs to be maximized is defined as follows
\begin{equation}\label{eq:19}
  L\left(\boldsymbol{\vartheta}\right)=\hat{\mathbb{E}}_{t}
  \left[\frac{\pi_{\boldsymbol{\vartheta}} \left(\boldsymbol{a}|\boldsymbol{s}\right)}
  {\pi_{\boldsymbol{\vartheta}_{\text{old}}} \left(\boldsymbol{a}|\boldsymbol{s}\right)}\hat{A}_{t}\right]
  =\hat{\mathbb{E}}_{t}\left[r_t\text{(}\boldsymbol{\vartheta}\text{)}\hat{A}_{t}\right]
\end{equation}
where $\boldsymbol{\vartheta}$ is the {NN} parameter, $r_t\text{(}\boldsymbol{\vartheta}\text{)}=\frac{\pi_{\boldsymbol{\vartheta}} \left(\boldsymbol{a}|\boldsymbol{s}\right)}{\pi_{\boldsymbol{\vartheta}_{\text{old}}} \left(\boldsymbol{a}|\boldsymbol{s}\right)}$ is the importance weight, $\hat{A}_{t}$ is the estimation of advantage function, and $\hat{\mathbb{E}}_{t}$ presents the estimation of expectation.
During the parameter update process, a batch of data is generated based on an existing policy $\pi_{\boldsymbol{\vartheta}_{\text{old}}}$ interacting with the environment, which is used to optimize the target policy $\pi_{\boldsymbol{\vartheta}}$. Batch sampling and batch processing of data are achieved by importance sampling and make agent easy to train. Excessive policy optimization leads {to difficulty in convergence} of the algorithm. In this paper, the PPO algorithm employs clipped surrogate objective to prevent excessive policy optimization
\begin{equation}\label{eq:20}
  \begin{split}
    L&^{CLIP}\left(\boldsymbol{\vartheta}\right)=\\
    &\hat{\mathbb{E}}_{t}\left[\min \left(r_t\text{(}\boldsymbol{\vartheta}\text{)}\hat{A}_{t},
    \mathrm{clip} \left(r_t\text{(}\boldsymbol{\vartheta}\text{)}, 1-\epsilon , 1+\epsilon \right)\hat{A}_{t}\right)\right]
  \end{split}
\end{equation}
where $\epsilon$ is a hyperparameter and $\mathrm{clip} \left(r_t\text{(}\boldsymbol{\vartheta}\text{)}, 1-\epsilon , 1+\epsilon \right)$ is the clipping function restricting the value of $r_t\text{(}\boldsymbol{\vartheta}\text{)}$ to the range $[1-\epsilon , 1+\epsilon]$.%, , $\epsilon=0.2$.

To solve the optimal flight control problem (\ref{eq:4}) in the DRL framework, two steps are involved in this paper: offline optimal flight control training and online adaptive weighting parameter tuning. More specifically, optimal state-feedback flight controllers represented as NNs for some representative arm length modes are first trained offline based on the PPO algorithm to get a set of optimal flight control laws. Then, an online weighting parameter tuning algorithm is proposed to obtain an overall flight control law by interpolation of the off-line trained optimal flight control laws for the morphing quadrotor with four variable-length arm rods.

%For the optimal state-feedback flight control design in the selected representative variable-arm-length mode, we are interested in training a policy function via the PPO algorithm and get an action vector $\boldsymbol{a}$ from the real-time flight state data vector $\boldsymbol{s}$, which guides the morphing quadrotor to fly along the preset path with better performance, i.e., designing a nonlinear state-feedback optimal controller $\boldsymbol{u}=\textit{\textbf{k}}(\boldsymbol{x})$ for the quadrotor \cite{chow2019lyapunov}, where $\boldsymbol{x}$ is the real-time flight state. Note that both state space and action space of the optimal quadrotor flight control are continuous, proximal policy optimization (PPO) algorithm \cite{schulman2017proximal} will be utilized to train the DRL-based optimal flight control scheme.

\section{Deep Reinforcement Learning for Offline Optimal Flight Control Design}
\label{sec:Deep Reinforcement Learning for Control Law Designing}
\subsection{Agent design}
\label{sec:Agent Design}
{Four rotors of the morphing quadrotor are chosen as actions of agent that is a $4$-dimensional action vector $\boldsymbol{a}$.} The environment with which the agent interacts is quadrotor dynamics and is modeled by a $12$-dimensional state vector $\boldsymbol{s}$ that is defined by (\ref{eq:6}). The agent makes a decision based on the observed state vector $\boldsymbol{s}$ and interacts with the environment through the action vector $\boldsymbol{a}$:
\begin{equation}\label{eq:7}
  \boldsymbol{a}=
  \begin{bmatrix}
    n_{1}\ n_{2}\ n_{3}\ n_{4}
  \end{bmatrix}^T
  \in \boldsymbol{\mathcal{A}}
\end{equation}
where $\boldsymbol{\mathcal{A}}$ is the action space, i.e., the set of all possible actions. The actions are angular velocities $n_{1}, n_{2}, n_{3}, n_{4} \in \left[0,n_{max}\right]$ of four rotors, and $n_{\text{max}}$ is the maximum rotor speed.

In order to enable the agent to extensively explore the action space with stable performance, we respectively adopt a stochastic policy in the train process and a deterministic policy in the test one. This policy is described by a probability density function, under which we will sample action vector randomly during training, and choose action vector with the largest probability in the course of testing. For the probability density function with the property that the action space is a finite domain, we resort to the Beta distribution with the definition domain $\left(0, 1\right)$ for each action dimension \cite{chou2017improving}. A finite domain action vector is obtained by sampling under the Beta distribution and multiplying by $n_{max}$. The corresponding probability density function of the Beta distribution is of the following form:
\begin{equation}\label{eq:8}
  f\left(x; \alpha, \beta\right)=\frac{1}{B\left(\alpha, \beta\right)}x^{\alpha-1}\left(1-x\right)^{\beta-1}
\end{equation}
where $B\left(\alpha, \beta\right)=\int_{0}^{1}x^{\alpha-1}\left(1-x\right)^{\beta-1}\, dx$ is the Beta function with $\alpha>0$ and $\beta > 0$ that are two parameters control the Beta distribution shape. {To facilitate the optimization of the agent policy, the probability density function has a bell curve with the value of $0$ at the boundary of the domain $\left(0, 1\right)$ similar to a normal distribution by choosing parameters $\alpha, \beta$ to be more than $1$.} The best policy for testing is to choose an action with the largest probability. An expectation of action $n_{mean}=\frac{\alpha}{\alpha+\beta}n_{max}$ is taken as a proxy for action with the largest probability to reduce the computational demand. %Although the expectation of the Beta distribution is different from the value with the largest probability, they are very close.

For a $4$-dimensional action vector $\boldsymbol{a}$, each component is described by an independent Beta distribution. So the policy $\pi(\boldsymbol{a}|\boldsymbol{s})$ can be written as a joint probability density function of the following form
\begin{equation}\label{eq:pi}
\begin{split}
  \pi(\boldsymbol{a}|\boldsymbol{s})
  &=\pi(n_{1}, n_{2}, n_{3}, n_{4}|\boldsymbol{s})\\
  &=\prod_{i=1}^{4}f\left(n_{i}; \alpha_{i}(\boldsymbol{s}), \beta_{i}(\boldsymbol{s})\right)
\end{split}
\end{equation}

\subsection{Neural network structure}
\label{sec:Networks Structure}
The agent includes an action network and a critic network, where an action network approximates the policy function and a critic network evaluates the policy. The inputs of these two NNs are {states} of the morphing quadrotor. According to discussions in the previous subsection, we have a $12$-dimensional state vector $\boldsymbol{s}$ and a $4$-dimensional action vector $\boldsymbol{a}$. Thus, the outputs of the policy function are probability density functions of Beta distribution describing the $4$-dimensional action vector, which can be fully described by two parameters $\alpha$ and $\beta$. As a result, the output layer of the action network has two terms: the parameter $\alpha$ and the one $\beta$. The output of the critic network is a scalar that describes the value of state vector in a given reward function.
\begin{figure}[!ht]
  \centering
  \subfloat[Actor Network]{\includegraphics[width=1\columnwidth]{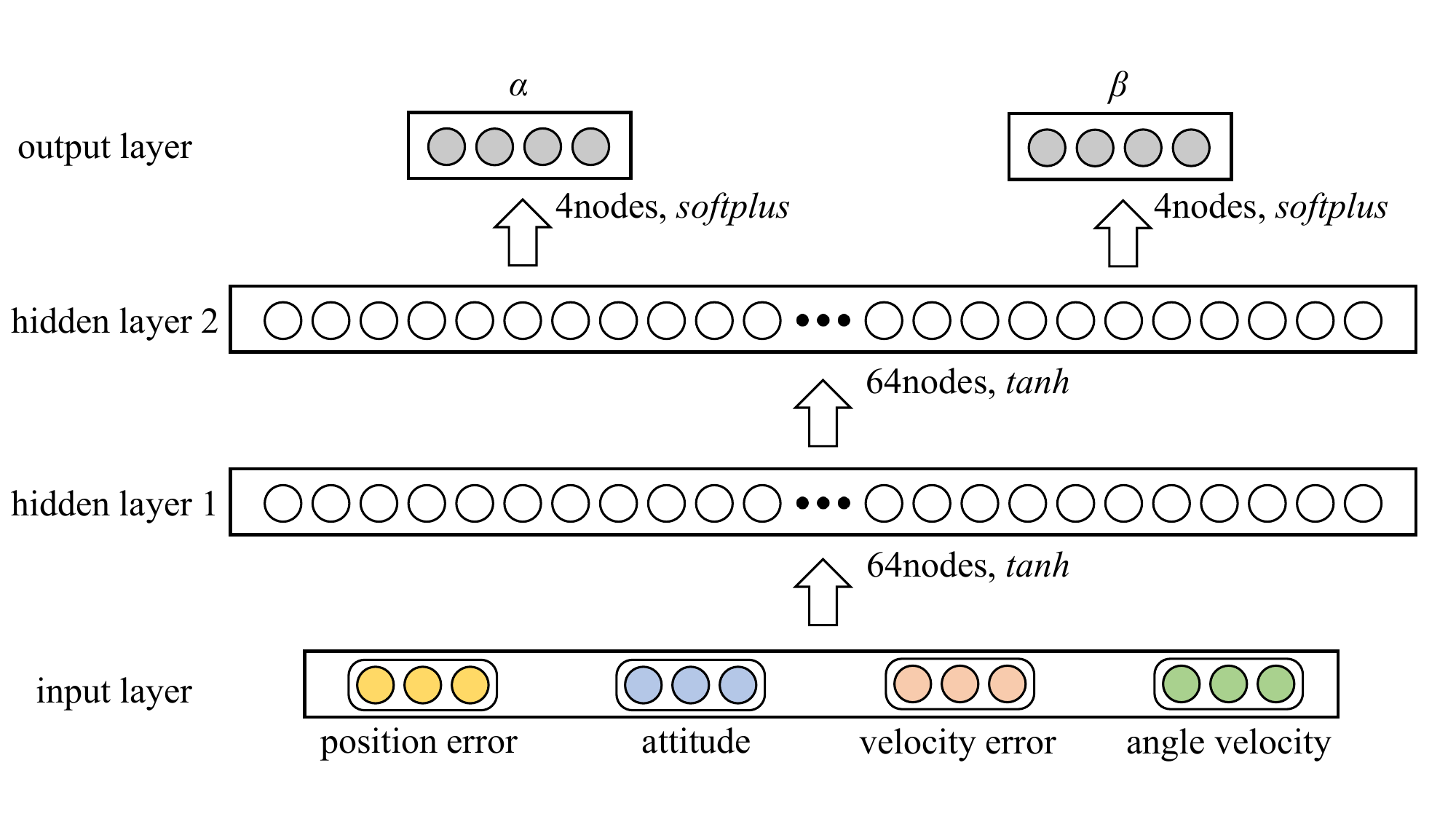}}\\
  %\vfill
  \subfloat[Critic Network]{\includegraphics[width=1\columnwidth]{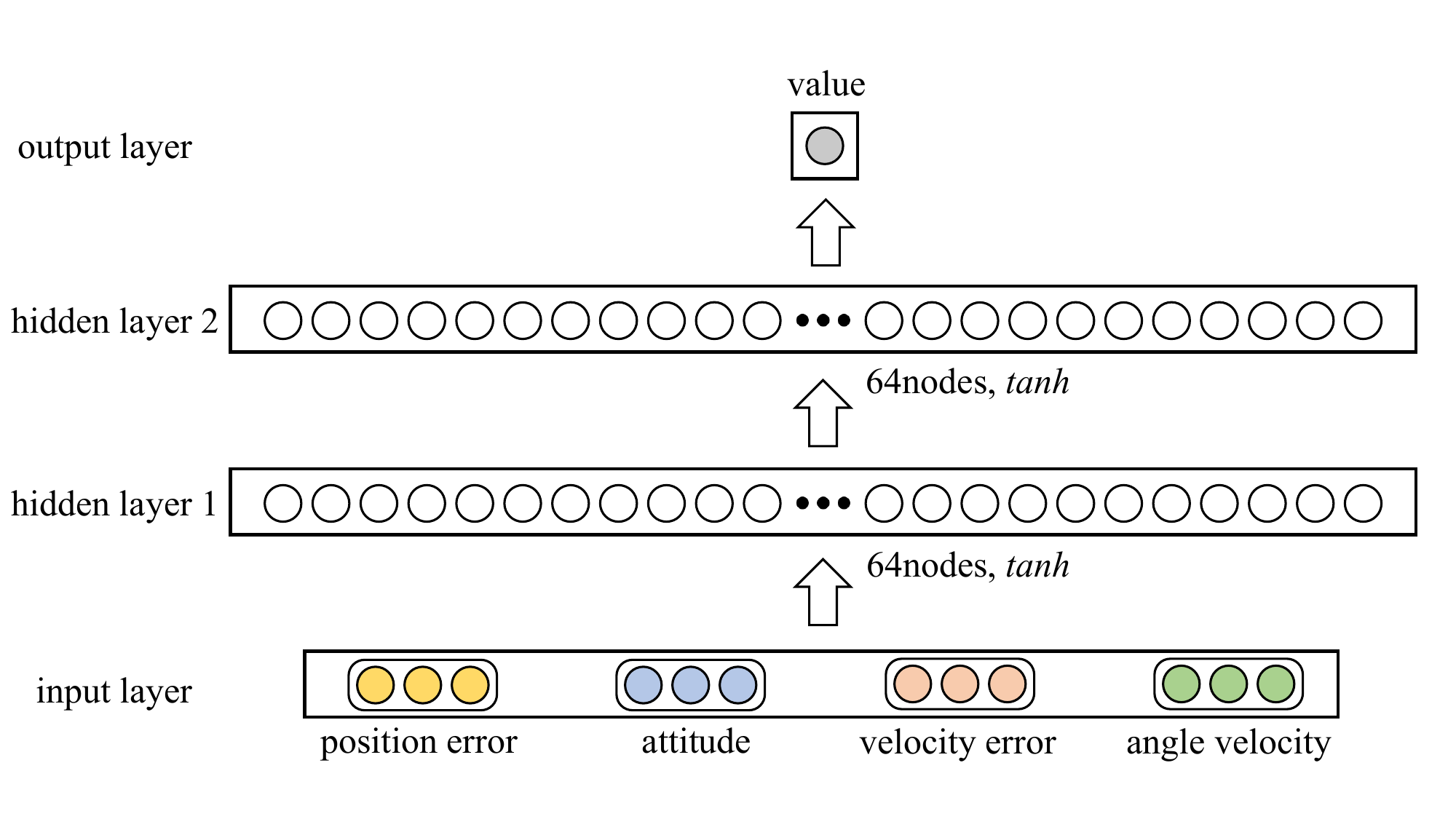}}
  \caption{The structure of networks.}
  \label{fig_3}
\end{figure}

The structure of both action and critic networks is shown in Fig. \ref{fig_3}. For these two networks, we use {fully connected NN} including two hidden layers, each with $64$ nodes, and the activation function is `tanh', respectively \cite{engstrom2020implementation}. We choose `softplus' as the activation function in the output layer of the action network and plus $1$ to ensure that the parameters $\alpha$ and $\beta$ of Beta distribution are more than $1$. The critic network's output is a scalar without any particular constraints, therefore we do not use any activation function for its output layer. The above-mentioned activation function expressions are respectively given by
\begin{equation}\label{eq:9}
  \tanh\left(x\right)=\frac{e^{x}-e^{-x}}{e^{x}+e^{-x}}
\end{equation}
\begin{equation}\label{eq:10}
  \mathrm{softplus}\left(x\right)=\ln\left(1+e^{x}\right)
\end{equation}

\subsection{Reward function}
\label{sec:Reward Function Design}
To get the simulation environment $Env\left(\boldsymbol{a}\right)$, we utilize the finite difference method to discrete the differential equation \eqref{eq:2}, where the {sampling period is set to be $\Delta T=0.1 \text{s}$}. Correspondingly, the performance metric \eqref{eq:5} is rewritten as a discrete form:
\begin{equation}\label{eq:5b}
  J=\sum_{k=0}^{T_{f}}F\left[\boldsymbol{s}\left(k\right), \boldsymbol{u}\left(k\right), k\right]
\end{equation}
where $T_{f}$ is the maximum number in the episode, and $F\left[\boldsymbol{s}\left(k\right), \boldsymbol{u}\left(k\right), k\right]$ is the terminal metric when $k=T_{f}$. The optimization objective in the optimal control problem \eqref{eq:4} is to minimize the performance metric $J$, while the aim of the DRL algorithm it to maximize the accumulated reward $G$
\begin{equation}\label{eq:5c}
  G=\sum_{k=0}^{T_{f}}\gamma^{k}r\left[\boldsymbol{s}\left(k\right), \boldsymbol{u}\left(k\right), k\right]
\end{equation}
where $0<\gamma<1$ is the discount factor and $r\left[\boldsymbol{s}\left(k\right), \boldsymbol{u}\left(k\right), k\right]$ is the reward at time $k$. In this situation, we choose $F\left[\boldsymbol{s}\left(k\right), \boldsymbol{u}\left(k\right), k\right]=-\gamma^{k}r\left[\boldsymbol{s}\left(k\right), \boldsymbol{u}\left(k\right), k\right]$ in the performance metric $J$. %The specific form of the reward $r\left[\boldsymbol{s}\left(k\right), \boldsymbol{u}\left(k\right), k\right]$ will be given later.}

%{In the process of interaction between the agent and the environment, a reward is given for each step. We will elaborate the reward function to realize the control objective \eqref{eq:4} and ensure the convergence of the algorithm.}
%We do not take the evolutions of states except position errors into consideration for the performance metric, and use a stochastic policy for training, which result in challenge for the algorithm.
{Rewards are added at each interaction step of transition process and at the end of each episode for terminal state. Standard Euclid norms of the position tracking error vector $\tilde{\textit{\textbf{x}}}$, the attitude angle vector $\bm{\varpi}$, the tracking error velocity vector $\dot{\tilde{\textit{\textbf{x}}}}$, the attitude angular velocity vector $\dot{\bm{\varpi}}$, and control input vector $\boldsymbol{u}$ are involved in the reward function $r$ with different weights. To make the exploration of agent more efficient}, penalty terms including the velocity error, attitude angle and attitude angular velocity are added into the reward function. At the beginning of the policy exploration, the quadrotor's position may deviate from the reference trajectory quickly. If the quadrotor's position deviation exceeds a certain value, this situation is regarded as the `crash' state. In this situation, the episode of training is immediately terminated with a high penalty and proceed into the next one directly {for the sake of saving computation overhead and blocking bad data for training}. A survival reward is added for policy optimization when the quadrotor is unable to successfully complete an episode. After the quadrotor is able to survive in an episode, the accumulated survival rewards are constant and no longer impact policy optimization. For the agent exploring different policies, we set a maximum time limit for each episode, and when the training reaches it, we end the episode and give an additional reward value based on terminal state.

According to above analysis, a reward function is designed as follows
\begin{equation}\label{eq:11}
\begin{split}
  r=-\mathrm{(}
    &c_{\tilde{\boldsymbol{x}}}\Vert\tilde{\boldsymbol{x}}\Vert+
    c_{\bm{\varpi}}\Vert\bm{\varpi}\Vert+
    c_{\dot{\tilde{\boldsymbol{x}}}}\Vert\dot{\tilde{\boldsymbol{x}}}\Vert+
    c_{\dot{\bm{\varpi}}}\Vert\dot{\bm{\varpi}}\Vert+ \\
    &c_{\boldsymbol{u}}\Vert\boldsymbol{u}\Vert+
    d_{e}c_{e}\Vert\tilde{\boldsymbol{x}}\Vert
    \mathrm{)}
    +d_{c}r_{c}+r_{t}
\end{split}
\end{equation}
where $\tilde{\boldsymbol{x}} \in \mathbb{R}^{3}$ is the position tracking error vector, $\bm{\varpi} \in \left[-\pi, \pi\right]^{3}$ is the attitude angle vector, $\dot{\tilde{\boldsymbol{x}}} \in \mathbb{R}^{3}$ is the tracking error velocity vector, $\dot{\bm{\varpi}} \in \mathbb{R}^{3}$ is the attitude angular velocity vector, $\boldsymbol{u} \in \left[0, n_{max}\right]^{4}$ is the control input vector, $r_{c} \in \mathbb{R}$ is a crash penalty, $c_{\tilde{\boldsymbol{x}}}, c_{\bm{\varpi}}, c_{\dot{\tilde{\boldsymbol{x}}}}, c_{\dot{\bm{\varpi}}}, c_{\boldsymbol{u}},c_{e} \in \mathbb{R}$ are the coefficients that adjust the importance among the various rewards, $r_{t} \in \mathbb{R}$ is a reward for survival of quadrotor, and $d_{c},d_{e} \in \{0,1\}$ are the {flags} of ending and defined by
\begin{equation}\label{eq:12}
  d_{c}=\begin{cases}
    1,&\text{if}\ \Vert\tilde{\boldsymbol{x}}\Vert>D;\\
    0,&\text{otherwise}.
  \end{cases}
\end{equation}
\begin{equation}\label{eq:13}
  d_{e}=\begin{cases}
    1,&\text{if}\ t=T_e;\\
    0,&\text{otherwise}.
  \end{cases}
\end{equation}
in which $D$ is the crash distance (when the tracking tracking error exceeds $D$, the episode ends as a crash), $t$ is the flight time, and $T_e$ is the set maximum time limit of the episode (when the maximum time limit is reached, the episode ends normally). Let $r_{\boldsymbol{s},\boldsymbol{u}}=(c_{\tilde{\boldsymbol{x}}}\Vert\tilde{\boldsymbol{x}}\Vert+c_{\bm{\varpi}}\Vert\bm{\varpi}\Vert+c_{\dot{\tilde{\boldsymbol{x}}}}\Vert\dot{\tilde{\boldsymbol{x}}}\Vert+c_{\dot{\bm{\varpi}}}\Vert\dot{\bm{\varpi}}\Vert+c_{\boldsymbol{u}}\Vert\boldsymbol{u}\Vert)$, a specific form of the reward function \eqref{eq:11} is chosen as follows
\begin{equation}\label{eq:20b}
  r=\begin{cases}
    -r_{\boldsymbol{s},\boldsymbol{u}}+r_{t},&k<T_{f};\\
    -r_{\boldsymbol{s},\boldsymbol{u}}+r_{c}+r_{t},&k=T_{f}<T_{e};\\
    -(r_{\boldsymbol{s},\boldsymbol{u}}+c_{e}\Vert\tilde{\boldsymbol{x}}\Vert)+r_{t},&k=T_{f}=T_{e}.
  \end{cases}
\end{equation}

\begin{remark} In fact, the deep NN training is divided into two stages. In the first stage, the agent learns from scratch to allow the quadrotor to successfully survive within an episode. This policy optimization is guided primarily by the survival reward $r_t$ and the crash penalty $r_c$. In the second stage, the agent optimizes the flight policy for a better flight performance. During this stage, the accumulated survival rewards are constant and the crash penalty is zero. This stage is mainly guided by the trajectory tracking error and the control inputs for policy optimization. %, where linear velocity error, attitude angle, attitude angular velocity, and terminal trajectory tracking error are chosen as penalty terms. %Although it is possible to divide training into two stages as described above, they are continuous while training, and even have the possibility to repeatedly switch between the two stages in the early phase of exploration.
 \end{remark}

\subsection{Loss Function}
\label{sec:Loss Function}
The critic network is updated based on the temporal difference (TD) error \cite{sutton1988learning}, of which the loss function is defined {as the mean square error (MSE) of $V_{\boldsymbol{s}}^{'}$ and $V_{\boldsymbol{s}}$}:
\begin{equation}\label{eq:21}
{L_{\text{Critic}}(\boldsymbol{\varphi})=\frac{1}{N}\left(V_{\boldsymbol{s}}^{'}-V_{\boldsymbol{s}}(\boldsymbol{\varphi})\right)}
\end{equation}
where $N$ is number of data in a batch, $V_{\boldsymbol{s}}^{'}=V_{\boldsymbol{s}}+A_{t}^{\gamma}$ is the TD-objective {and $V_{\boldsymbol{s}}$ is the value of state ${\boldsymbol{s}}$}.
 For a better tradeoff between bias and variance in the value function estimation, the TD($\lambda$) algorithm is used, and $A_{t}^{\gamma}$ is the generalized advantage estimation (GAE) \cite{schulman2015high}:
\begin{equation}\label{eq:22}
  \begin{split}
    A_{t}^{\gamma}&=\left(1-\lambda\right)\left[A_{t}^{(1)}+\lambda A_{t}^{(2)}+\lambda^{2} A_{t}^{(3)}+\cdots \right]\\
  &=\sum_{n=0}^{\infty}\left(\gamma \lambda\right)^{n}\delta_{t+n}
  \end{split}
\end{equation}
where $A_{t}^{(n)}=\sum_{i=0}^{n}\gamma^{i}\delta_{t+i}$ is the sum of $n$-step TD errors and $\delta_{t}=r_t+\gamma V_{\boldsymbol{s},t+1}-V_{\boldsymbol{s},t}$ is the TD-error.

As the maximum number in the episode is $T_{f}$, we know $\delta_{t+n}=0$, $\forall(t+n)>T_f$ and the expression \eqref{eq:22} is simplified as
\begin{equation}\label{eq:22b}
  \begin{split}
    A_{t}^{\gamma}=\sum_{n=0}^{T_{f}-t}\left(\gamma \lambda\right)^{n}\delta_{t+n}
  \end{split}
\end{equation}

The action network is updated with the clipped surrogate objective, in which the loss function is defined as follows:
\begin{equation}\label{eq:23}
  \begin{split}
    &L_{\text{Actor}}=-\hat{\mathbb{E}}_{t}\\
    &\left[\min \text{(}r_t\text{(}\boldsymbol{\vartheta}\text{)}\hat{A}_{t},
    \mathrm{clip} \text{(}r_t\text{(}\boldsymbol{\vartheta}\text{)}, 1-\epsilon , 1+\epsilon \text{)}\hat{A}_{t}\text{)}+cH\left[\pi_{\boldsymbol{\vartheta}}\text{(}\boldsymbol{a}_{t}|\boldsymbol{s}_{t}\text{)}\right]\right]
  \end{split}
\end{equation}
where $c>0$ is the coefficient of policy entropy and $H\left[\pi_{\boldsymbol{\vartheta}}\text{(}\boldsymbol{a}_{t}|\boldsymbol{s}_{t}\text{)}\right]$ is the policy entropy. For the collected discrete data, the policy entropy can be expressed as
\begin{equation}\label{eq:23b}
  H\left[\pi_{\boldsymbol{\vartheta}}\text{(}\boldsymbol{a}_{t}|\boldsymbol{s}_{t}\text{)}\right]
  =-\sum_{\boldsymbol{a}_{t}} \pi_{\boldsymbol{\vartheta}}\text{(}\boldsymbol{a}_{t}|\boldsymbol{s}_{t}\text{)}\log\pi_{\boldsymbol{\vartheta}}\text{(}\boldsymbol{a}_{t}|\boldsymbol{s}_{t}\text{)}
\end{equation}

The introduction of policy entropy regularization allows the policy to be optimized in a much more random way and enhances the agent's explore ability of the action space.

\subsection{Updating Process}
\label{sec:Updating Process}
The agent collects data and updates the network via the PPO algorithm while interacting with the environment. A \textit{ReplayBuffer} is set to store the data of interaction, including state $\boldsymbol{s}$, action $\boldsymbol{a}_{0}$, probability $p_{\pi}\left(\boldsymbol{a}_{0}\right)$, reward $r$, next state $\boldsymbol{s}'$, and {flag $d$}. Whenever the data in the \textit{ReplayBuffer} is full, the agent performs a task of networks update including $k$ epochs, and empties \textit{ReplayBuffer} to restart storing the data. For each epoch, the data is divided into a number of mini-batchsizes randomly and the \textit{Adam optimizer} is used to update networks' weights. During training, the following tricks are used to improve the performance of the proposed optimal flight control scheme:
\begin{figure}[!t]
  \centering
  \includegraphics[width=1\columnwidth]{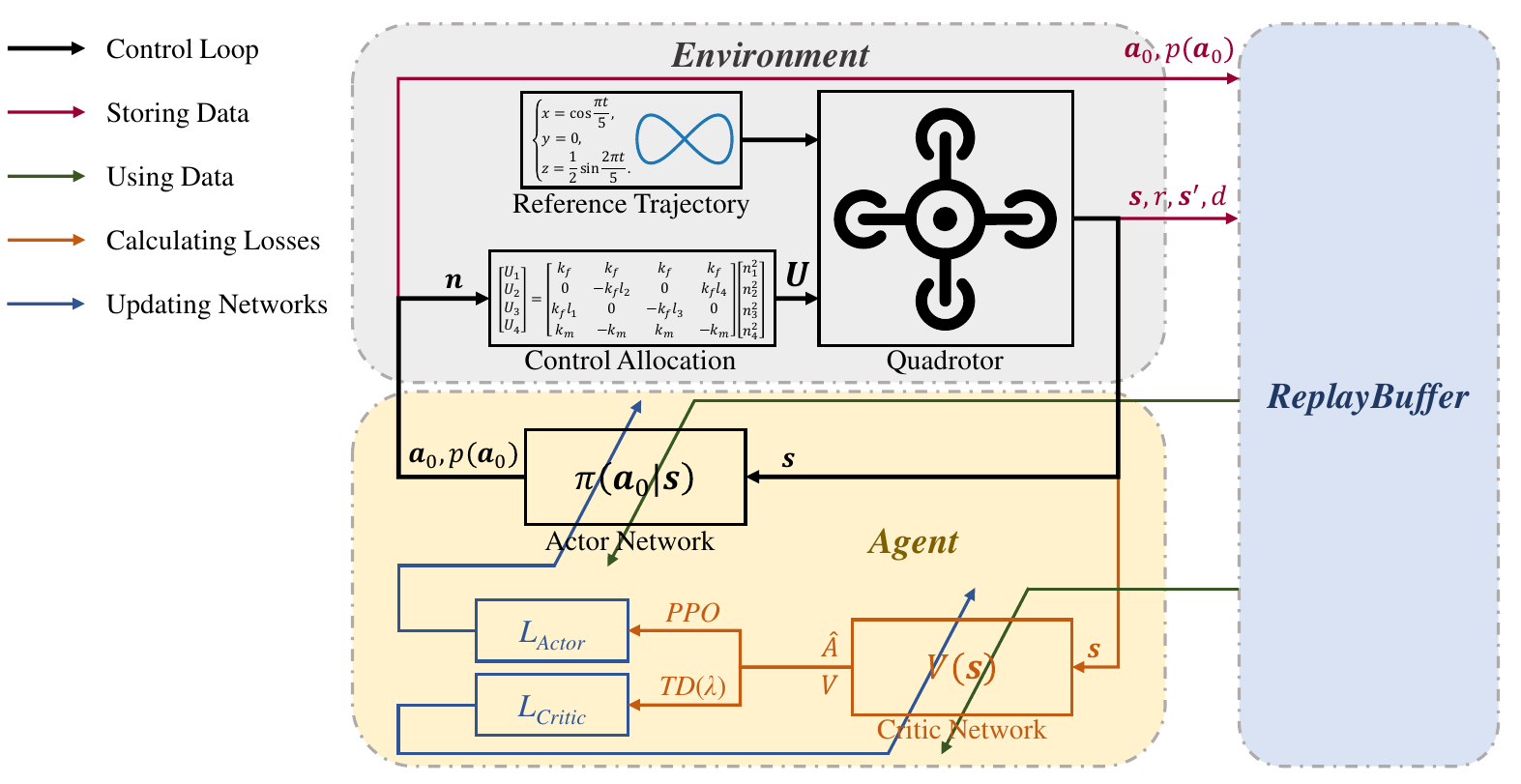}
  \caption{The PPO algorithm. There are three parts: \textit{Environment}, \textit{Agent}, and \textit{ReplayBuffer}.
  \textit{Environment} is quadrotor dynamics and is used for interaction to generate states;
  \textit{Agent} includes an action network and an evaluation network and is used for state evaluation and policy learning;
   and \textit{ReplayBuffer} is used to store interaction data.}
  \label{PPO_Alg}
\end{figure}
\begin{algorithm}[!t]
  \caption{PPO-based optimal flight control in a specific arm length mode}\label{alg:alg1}
  \textbf{Input:} The reference trajectory $\boldsymbol{x}_r$\\
  \textbf{Hyperparameter:} Entropy coefficient $c$, clip parameter $\epsilon$, motor maximum velocity $n_{max}$, discount factor $\gamma$, parameter of $\lambda$-return $\lambda$, learning rate of actor network $\eta_{a}$ and critic network $\eta_{c}$\\
  \textbf{Require:} Quadrotor dynamics environment $Env\left(\boldsymbol{a}\right)$:$\boldsymbol{a}\mapsto \boldsymbol{s}', r, d$\\
  \textbf{Initialize:} iteration $= 0$, count $= 0$, \textit{ReplayBuffer}, environment $Env\left(\boldsymbol{a}\right)$,
  actor network $\pi_{\boldsymbol{\vartheta}} \left(\boldsymbol{a}|\boldsymbol{s}\right)$, critic network $V_{\boldsymbol{\varphi}} \left(\boldsymbol{s}\right)$, optimizer \textit{Adam}, and arm lengths $\boldsymbol{l}=\boldsymbol{l}_{\text{set}}$\\
  \textbf{Result:} Trained $\pi_{\boldsymbol{\vartheta}} \left(\boldsymbol{a}|\boldsymbol{s}\right)$ and $V_{\boldsymbol{\varphi}} \left(\boldsymbol{s}\right)$
  \begin{algorithmic}[1]
    \While{iter $< T$} \Comment{Training $T$ steps}
      \State Reset $Env\left(\boldsymbol{a}\right)$ with $\boldsymbol{x}_r$ and get return $\boldsymbol{s}, d$
      \While{$d = 0$} \Comment{interacting in an episode}
        \State Sample $\boldsymbol{a}_{0}$ under $\pi_{\boldsymbol{\vartheta}} \left(\boldsymbol{a}|\boldsymbol{s}\right)$ and get $p_{\pi}\left(\boldsymbol{a}_{0}\right)$
        \State Compute action $\boldsymbol{a}=\boldsymbol{a}_{0}\times n_{max}$
        \State Interact with $Env\left(\boldsymbol{a}\right)$ by $\boldsymbol{a}$ and get returns $\boldsymbol{s}', r, d$
        \State store $\{\boldsymbol{s}, \boldsymbol{a}_{0}, p_{\pi}\left(\boldsymbol{a}_{0}\right), r, \boldsymbol{s}', d\}$ in \textit{ReplayBuffer}
        \State $\boldsymbol{s}\leftarrow \boldsymbol{s}'$, count $\leftarrow$ count $+ 1$, iter $\leftarrow$ iter $+ 1$
        \If{count $= N$} \Comment{Updating}
          \State Compute $V_{\boldsymbol{s}}$ and $V_{\boldsymbol{s}}^{'}$ of $\boldsymbol{s}$ and $\boldsymbol{s}'$ by $V_{\boldsymbol{\varphi}} \left(\boldsymbol{s}\right)$
          \State Compute GAE by Eq.\eqref{eq:22}
          \For{$i=1, 2, \cdots , K$}
            \State Randomly split data into $M$ minibatches
            \For{$j=1, 2, \cdots , M$}
              \State Compute $L\text{(}\boldsymbol{\vartheta}\text{)}$ by Eq.\eqref{eq:19}
              \State Updating $\pi_{\boldsymbol{\vartheta}} \left(\boldsymbol{a}|\boldsymbol{s}\right)$ using Eq.\eqref{eq:23}
              \State Updating $V_{\boldsymbol{\varphi}} \left(\boldsymbol{s}\right)$ using Eq.\eqref{eq:21}
            \EndFor
          \EndFor
          \State Empty \textit{ReplayBuffer}, count $\leftarrow 0$
        \EndIf
      \EndWhile
    \EndWhile
  \end{algorithmic}
  \label{alg1}
\end{algorithm}
\begin{itemize}
   \item Orthogonal initialization is used for the networks' weights to prevent problems such as gradient vanishing and gradient explosion at the beginning of training.
   \item Advantage normalization is used in each batchsize \cite{tucker2018mirage}.
   \item Reward scaling is used for each reward \cite{engstrom2020implementation}.
   \item Linear decay learning rate is used in the Adam optimizer \cite{zhang2018improved}, \cite{liu2019variance}.
   \item Excessively large gradients are clipped before optimization \cite{zhang2019gradient}.
 \end{itemize}
 The algorithm details are shown in \textbf{Algorithm 1} and Fig. \ref{PPO_Alg}. By repeatedly applying \textbf{Algorithm 1} for the selected representative length modes of arm rods, the corresponding DRL-based offline optimal flight control scheme can be obtained.

\section{Combined Deep Reinforcement Learning Flight Control via Weighting Combination}
\label{sec:Online Adaptive Control Law}
%, where the corresponding weighting values can be obtained by resorting to the online adaptive parameter estimation.
%\subsection{cc-DRL flight control via weighting combination}\label{sec:Combination of Control laws}
For arbitrary lengths of four quadrotor arm rods, via convex combination, they can be represented as a linear combination of some selected representative arm length modes. In the light of this fact, a cc-DRL flight control scheme can be obtained by interpolation of the optimal flight control schemes that are trained offline for the representative arm length modes. In this way, a cc-DRL flight control law $\boldsymbol{u}_{\text{cc-DRL}}$ is directly obtained from a set of trained optimal flight control laws $\boldsymbol{\mathcal{U}}=\{\boldsymbol{u}_i|\boldsymbol{u}_i=\pi_{\boldsymbol{\vartheta}, i} \left(\boldsymbol{a}|\boldsymbol{s}\right)\}$, i.e.,
\begin{equation}\label{eq:24}
  \begin{split}
    \boldsymbol{u}_{\text{cc-DRL}}=\sum_{i=1}^{n}\chi_{i} \boldsymbol{u}_{i}
  \end{split}
\end{equation}
where $\boldsymbol{u}_{i} \in \mathbb{R}^{4}$, $i\in\{1,2,\cdots,n\}$ are the offline trained optimal flight control laws for the $n$ selected representative arm lengths $\boldsymbol{l}_{i} \in \mathbb{R}^{4}$, $i\in\{1,2,\cdots,n\}$ and $\chi_i$, $i\in\{1,2,\cdots,n\}$ are the combination weight values satisfying
\begin{equation}\label{eq:25a}
\boldsymbol{l}(t)=\sum_{i=1}^{n}\chi_{i}\boldsymbol{l}_{i},
\end{equation}
in which $\boldsymbol{l}(t) \in \mathbb{R}^{4}$ is the arbitrary current length vector of four quadrotor arm rods.

Assume that the length range of each arm is set to be $[l_{\text{min}}, l_{\text{max}}]$, and an arbitrary arm length vector $\boldsymbol{l}={\left[l_{1}\ l_{2}\ l_{3}\ l_{4}\right]}\in\mathbb{R}^4$ is chosen from the set $[l_{\min},l_{\max}]^{4}$, i.e., $\boldsymbol{l}\in \boldsymbol{\mathcal{C}}\triangleq\{\boldsymbol{l}|\boldsymbol{l} \in[l_{\min},l_{\max}]^{4} \} \subset \mathbb{R}^4$. Obviously, $\boldsymbol{\mathcal{C}}$ is a convex set, which is a hypercube with $16$ vertices. At these $16$ vertices, arm length vectors are selected as representative modes. That is, the positive integer $n$ in \eqref{eq:24} and \eqref{eq:25a} is $16$, i.e., $n=16$.
%\subsection{Arbitrary arm length represented by convex combination}\label{sec:Convex Combination of Arbitrary Arm Length}
%Let $[l_{\min},l_{\max}]$ be the length range of each arm rod and $\boldsymbol{l}={\left[l_{1}\ l_{2}\ l_{3}\ l_{4}\right]}\in\mathbb{R}^4$ be arbitrary lengths of four arm rods in the morphing quadrotor. Obviously, $\boldsymbol{l}\in \boldsymbol{\mathcal{C}}=\{\boldsymbol{l}|\boldsymbol{l} \in[l_{\min},l_{\max}]^{4} \} \subset \mathbb{R}^4$ and the convex set $\boldsymbol{\mathcal{C}}$ is a hypercube with 16 vertices. Via the convex combination technique, arbitrary length $\boldsymbol{l}_{\text{tar}}$ of four arm rods can be represented as follows
%\begin{equation}\label{eq:25a}
%  \sum_{i=1}^{16}\chi_{i}\boldsymbol{l}_{i}=\boldsymbol{l}_{\text{tar}}
%\end{equation}
%where $\chi_{i}$ and $\boldsymbol{l}_{i}$, $i\in\{1,2,\cdots,16\}$ are weighting parameters to be determined and 16 selected representative arm length modes, respectively.

{The minimum norm solution to Eq. \eqref{eq:25a} can be easily solved by right pseudo-reverse, but we want to obtain its the maximum norm solution.} To do this, by Caratheodory's theorem \cite{eckhoff1993helly}, any element in a convex set $\boldsymbol{\mathcal{C}}$ in $\mathbb{R}^{4}$ can be represented by a convex combination of $5$ or fewer vertices. The maximum norm solution to Eq.\eqref{eq:25a} can be formulated as the following non-convex quadratic programming (NCQP) problem:
\begin{align}\label{eq:25b}
  &\min-\sum_{i=1}^{n}\chi_{i}^{2},\nonumber\\
  &\quad \text{s.t.}\ \begin{cases}
    \sum_{i=1}^{n}\chi_{i}=1,\\
    \sum_{i=1}^{n}\chi_{i}\boldsymbol{l}_{i}=\boldsymbol{l}_{\text{tar}},\\
    \chi_{i} \geqslant  0, i \in \{1, 2, \cdots, n\}.
  \end{cases}
\end{align}

Generally, it is difficult to obtain an analytical solution directly to the problem (\ref{eq:25b}). The Sequential Least Squares Programming (SLSQP) algorithm will be used to solve iteratively \cite{kraft1988software}. In order to obtain a linear combination with as few representative arm length modes as possible, during the iterations, if the solution contains more than $5$ nonzero values, we will resolve the problem until the nonzero values are less than or equal to $5$ and normalize the solution. The details of the algorithm is shown in \textbf{Algorithm 2}.
Although the proposed algorithm may fall into a local optimum, the subsequent use of the solution for a linear combination of control laws only results in a small difference in performance compared to the global optimum. This issue is far less significant than the effect of randomness in the DRL algorithm.% Therefore, in order to save computational overhead making the algorithm deployable online, some of the performance can be sacrificed to use a local optimum solution which satisfies the conditions.

\begin{algorithm}[!t]
\caption{NCQP for Combination Coefficients}\label{alg:alg2}
\textbf{Input:} The target arm lengths $\boldsymbol{l}_{\text{tar}}$\\
\textbf{Require:} SLSQP: \{NCQP, $\boldsymbol{\chi}_{0}$\} $\to \boldsymbol{\chi}=\left[\chi_{1}, \chi_{2}, \cdots , \chi_{16}\right]$\\
\textbf{Initialize:} num $= 16$\\
\textbf{Ensure:} The {NCQP} problem Eq.\eqref{eq:25b}\\
\textbf{Result:} Solution of programming $\boldsymbol{\chi}$
\begin{algorithmic}[1]
\While{num $> 5$}
  \State Randomly generated $\boldsymbol{\chi}_{0}$ with normalization
  \State Solving Eq.\eqref{eq:25b} using SLSQP with $\boldsymbol{\chi}_{0}$ as initial value
  \State num $\leftarrow$ Number of nonzero elements of $\boldsymbol{\chi}$
  \EndWhile
\end{algorithmic}
\label{alg2}
\end{algorithm}
\begin{algorithm}[!b]
  \caption{cc-DRL Flight Control Law via Online Weighting Combination}\label{alg:alg3}
  \textbf{Input:} The reference trajectory $\boldsymbol{x}_r$\\
  \textbf{Require:} A trained set $\boldsymbol{\mathcal{U}}=\{\boldsymbol{u}_i|\boldsymbol{u}_i=\pi_{\boldsymbol{\vartheta}, i} \left(\boldsymbol{a}|\boldsymbol{s}\right)\}$, and quadrotor dynamics environment $Env\left(\boldsymbol{a}\right)$:$\boldsymbol{a}\mapsto \boldsymbol{s}', r, d$\\
  \textbf{Initialize:} Environment $Env\left(\boldsymbol{a}\right)$, and arm lengths {$\boldsymbol{l}=[0.15\ 0.15\ 0.15\ 0.15]$}\\
  \textbf{Output:} The state sequence $\{\boldsymbol{s}\}$
  \begin{algorithmic}[1]
  \State Reset $Env\left(\boldsymbol{a}\right)$ and get return $\boldsymbol{s}, d$
  \While{$d = 0$}
    \State Get $\boldsymbol{l}_{\text{new}}$ from external command
    \If{$\boldsymbol{l}\neq \boldsymbol{l}_{\text{new}}$}
      \State $\boldsymbol{l}\leftarrow \boldsymbol{l}_{\text{new}}$
      \State Compute $\boldsymbol{\chi}$ by \textbf{Algorithm 2} with input $\boldsymbol{l}$
    \EndIf
    \For{$i=1, 2, \cdots, 16$}
      \If{$\chi_{i} \neq 0$}
      \State Compute expectation of $\pi_{\boldsymbol{\vartheta}, i} \left(\boldsymbol{a}|\boldsymbol{s}\right)$ as $\boldsymbol{a}_{\text{mean}, i}$
      \State Compute $\boldsymbol{a}_{i}=\boldsymbol{a}_{\text{mean}, i}\times n_{\text{max}}$
      \Else
      \State $\boldsymbol{a}_{i}=0$
      \EndIf
    \EndFor
    \State Compute action  $\boldsymbol{a}=\sum_{i=1}^{16}\chi_{i}\boldsymbol{a}_{i}$
    \State Interact with $Env\left(\boldsymbol{a}\right)$ by $\boldsymbol{a}$ and get returns $\boldsymbol{s}', r, d$
    \State $\boldsymbol{s}\leftarrow \boldsymbol{s}'$
  \EndWhile
  \end{algorithmic}
  \label{alg3}
  \end{algorithm}
\begin{remark}  \textbf{Algorithm 2} can compute online. To further improve the online computation speed and save resource overhead, a little computational accuracy can be discarded and a NN can be trained offline to describe the relationship between arm lengths and coefficients.
\end{remark}

%\subsection{Online weighting coefficient estimation}\label{sec:Online Adaptive Convex Combination}
The arm length variation of morphing quadrotor is ruled by an external command according to the environment change or the task execution requirement. In this paper, we only consider the control effect of quadrotor flight dynamics of the morphing quadrotor. When the arm length variation command is active, the variation of arm lengths is a slow process compared to the quadrotor dynamics. The arm length variation command is simulated by a ramp input instead of a step input. Hence, we assume that the arm lengths are available in real time and neglects the error between the actual lengths and their reference signals. A cc-DRL flight control law is obtained via \textbf{Algorithm 3} from the offline trained optimal flight control laws.

\section{Simulation Study}
\label{sec:Simulation}
\subsection{Simulation environment settings}
\label{sec:Parameters Design}
\begin{figure}[!ht]
  \centering
  \includegraphics[width=1\columnwidth]{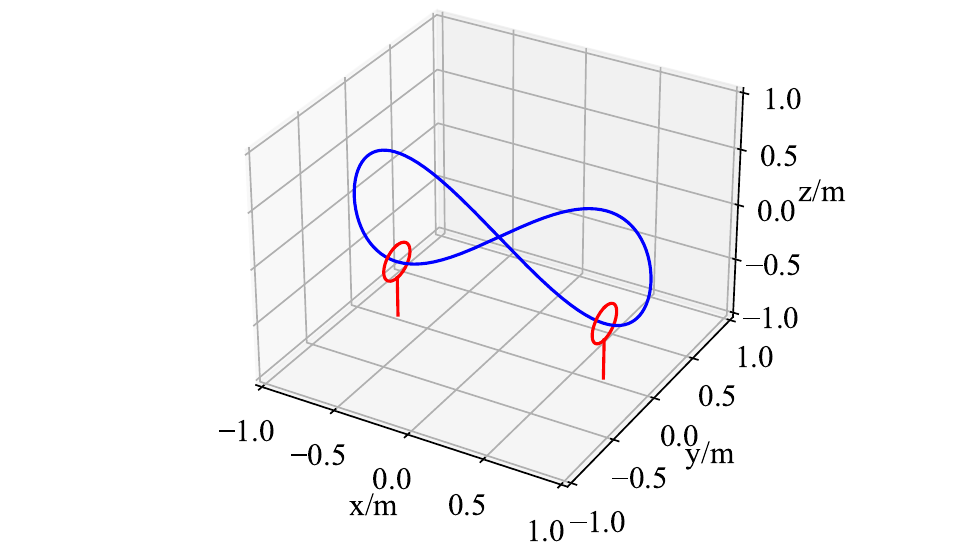}
  \caption{Figure-8 flight trajectory.}
  \label{fig_4}
\end{figure}

%We assume that the active variations of the arm lengths of quadrotor to adapt to different environments or tasks are caused by an event-triggered mechanism rather than a control signal given by our algorithm, e.g., the quadrotor is going to through a narrow space, which triggers the command to shorten the arm lengths according to sensors or human beings. What we are interested in is the better flight performance of the quadrotor in arbitrary variant modes or during variant transitions.
A small morphing quadrotor is discussed in this section, whose parameters and their chosen values are shown in TABLE I. Due to the inertia moments are influenced by the arm length changing, TABLE I only gives the value of inertia moments of the morphing quadrotor with a shortest arm length. The length range of each am rod is $[0.15, 0.25]$m and the upper limit of rotor speed is set to be $1000$ r/min, i.e., $l_{\min}=0.15$m, $l_{\max}=0.25$m, and $n_{max}=1000 \text{r/min}$. The reference flight trajectory is ruled by
\begin{equation}\label{eq:26}
  \begin{cases}
    x\left(t\right)=\cos \frac{\pi t}{5},\\
    y\left(t\right)=0,\\
    z\left(t\right)=\frac{1}{2}\sin \frac{2 \pi t}{5},
  \end{cases}t \in \left[0, 20\right].
\end{equation}
which is a figure-8 flight trajectory in the $xOz$ plane as shown in Fig. \ref{fig_4} and is a commonly used control benchmark \cite{o2022neural}\footnote{Of course, the other reference flight trajectories can also be used to test the performance of the proposed online flight control scheme}. At each episode for a total of $20$ second, the quadrotor completes the flight task for two circles.

\begin{table}[!htbp]
\centering
\label{tab:1}
\caption{Parameters of quadrotor}
  \begin{tabular}{cll}
  \toprule
  Notation &Description &Value\\
  \midrule
  $m$ &Mass of quadrotor &$1.732$kg\\
  $I_{x}$ &Moment of inertia about $X$-axis &$0.0375$kg$\cdot$m$^2$\\
  $I_{y}$ &Moment of inertia about $Y$-axis &$0.0375$kg$\cdot$m$^2$\\
  $I_{z}$ &Moment of inertia about $Z$-axis &$0.0749$kg$\cdot$m$^2$\\
  $k_f$ &Coefficient of rotor lifting force &$3.03\times 10^{-5}$N/rad$^2$\\
  $k_m$ &Coefficient of motor anti-torque &$5.5\times 10^{-5}$N$\cdot$m/rad$^2$\\
  $n_{\text{max}}$ &Maximum speed of motor &$1000$rpm\\
  $\Delta T$ &Sampling interval &$0.1$s\\
  \bottomrule
  \end{tabular}
\end{table}

\begin{table}[!htbp]
  \centering
  \label{tab:2}
  \caption{Parameters of Reward Function}
  \begin{tabular}{cll}
    \toprule
    Notation &Description &Value\\
    \midrule
    $c_{\tilde{\boldsymbol{x}}}$ &Coefficient of trajectory error &$4\times 10^{-1}$\\
    $c_{\boldsymbol{\theta}}$ &Coefficient of attitude angle &$2\times 10^{-2}$\\
    $c_{\dot{\tilde{\boldsymbol{x}}}}$ &Coefficient of trajectory velocity error &$3\times 10^{-2}$\\
    $c_{\dot{\boldsymbol{\theta}}}$ &Coefficient of attitude angular velocity &$5\times 10^{-2}$\\
    $c_{\boldsymbol{u}}$ &Coefficient of control input &$1\times 10^{-4}$\\
    $c_{e}$ &Coefficient of terminal trajectory error  &$10$\\
    $r_{c}$ &Penalty of crash &$-150$\\
    $r_{t}$ &Reward of survival &$1$\\
    $D$ &Boundary of crash &$5$m\\
    $T_{e}$ &Maximum number of steps per episode &$200$\\
    \bottomrule
  \end{tabular}
\end{table}

The parameters of reward function for DRL are shown in TABLE II. The trained DRL control law should improve the trajectory tracking performance and save energy consumption while maintaining the given tracking accuracy. Hence, both trajectory error and control inputs are the two items occupying a larger proportion in the reward function. Penalty terms for linear velocity error, attitude angle, and attitude angular velocity are added with a smaller proportion to ensure and accelerate the training convergence. Without considering the additional rewards, {the term of control inputs in the reward function is second only to the trajectory error one.} Otherwise, to achieve excellent convergence performance, the model is trained for $5\times 10^{7}$ steps and updated via \textit{Adam} optimizer with parameter $\epsilon_{Adam}=1\times 10^{-5}$. The details of the algorithm parameter values are shown in TABLE III.
 \begin{table}[!bt]
  \centering
  \label{tab:3}
  \caption{Parameters of PPO Algorithm}
  \begin{tabular}{cll}
    \toprule
    Notation &Description &Value\\
    \midrule
    $T$ &Maximum training steps &$5\times 10^{7}$\\
    $N$ &Maximum capacity of \textit{ReplayBuffer}&$2048$\\
    $K$ &Number of epochs for each update&$10$\\
    $c$ &Coefficient of policy entropy &$0.01$\\
    $\epsilon$ &Parameter of clip &$0.2$\\
    $\gamma$ &Discount factor &$0.99$\\
    $\lambda$ &Parameter of $\lambda$-return &$0.95$\\
    $\eta_{a}$ &Learning rate of actor network &$3\times 10^{-5}$\\
    $\eta_{c}$ &Learning rate of critic network &$3\times 10^{-5}$\\
    $\epsilon_{Adam}$ &Parameter of Adam &$1\times 10^{-5}$\\
    \bottomrule
  \end{tabular}
\end{table}

\subsection{DRL-based offline optimal flight control design}
\label{sec:PPO Algorithm Performance}
\begin{figure}[!ht]
  \centering
  \includegraphics[width=0.7\columnwidth]{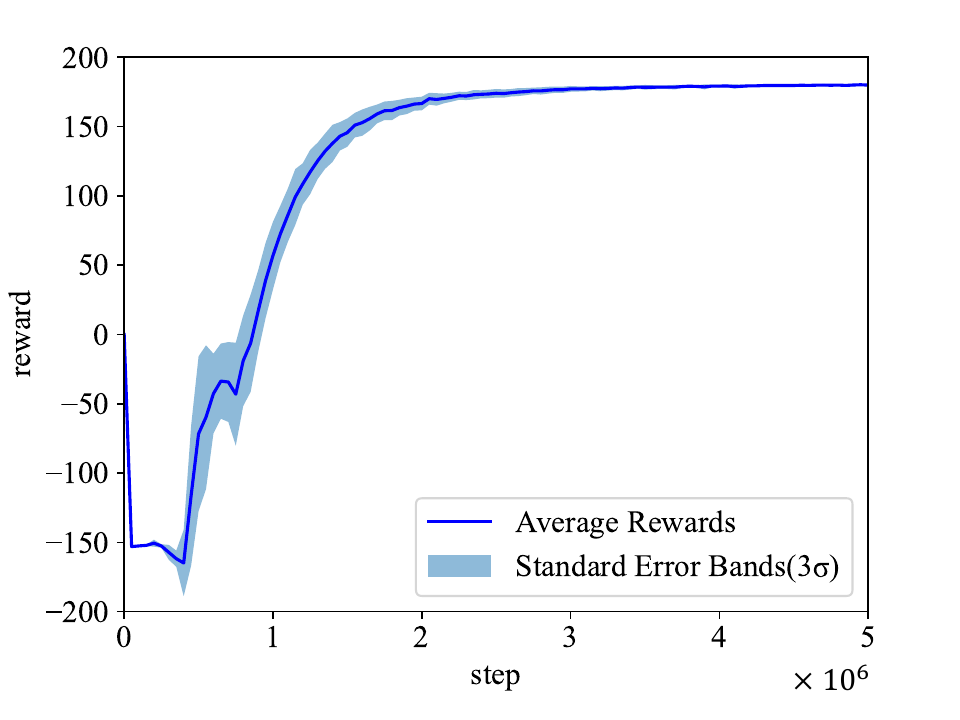}
  \caption{The averaged reward-curve for $16$ selected length modes for four arm rods}
  \label{fig_5}
\end{figure}

For any length vector $\boldsymbol{l}={\left[l_{1}\ l_{2}\ l_{3}\ l_{4}\right]}\in\mathbb{R}^4$ of four arm rods in the morphing quadrotor, they changes in the convex set $\boldsymbol{\mathcal{C}}$ that is a hypercube with $16$ vertices, i.e., $\boldsymbol{l}\in \boldsymbol{\mathcal{C}}=\{\boldsymbol{l}|\boldsymbol{l} \in[l_{\min},l_{\max}]^{4} \} \subset \mathbb{R}^4$. %\begin{equation}\label{eq:25a}
%  \sum_{i=1}^{16}\chi_{i}\boldsymbol{l}_{i}=\boldsymbol{l}_{\text{tar}}
%\end{equation}
%where $\chi_{i}$ and $\boldsymbol{l}_{i}$, $i\in\{1,2,\cdots,16\}$ are weighting parameters to be determined and 16 selected representative arm length modes, respectively.
Hence, $16$ length modes for four arm rods are selected in TABLE IV, where ``1" is used to represent an arm length of $0.25$m and ``0" represents an arm length of $0.15$m. By \textbf{Algorithm 1}, the final training rewards of each mode are also shown in TABLE IV and {the reward-curve of the 16 selected length modes is shown in Fig. \ref{fig_5}} for four arm rods. As shown in Fig. \ref{fig_5}, the reward is negative at the beginning of the training. This is because the agent is unable to successfully complete trajectory tracking task within an episode and thus receives a negative cumulative reward. With the increment of training steps, the agent gradually explores an action policy that can guide the quadrotor to complete trajectory tracking task within an episode. On the basis of this action policy, the agent further explores the optimal action policy and the accumulated reward gradually raises. Over a long period, the accumulated reward rises slowly within a gradual weakening of its oscillation, and the agent is fine-tuning its policy.
\begin{table}[!h]
  \centering
  \label{tab:4}
  \caption{Rewards of 16 selected length modes for four arm rods}
  \begin{tabular}{cccccc}
    \toprule
    \multirow{2}*{Mode} &\multicolumn{4}{c}{Arm length/m} &\multirow{2}*{Rewards}\\
    \cline{2-5}
    &$L_{1}$ &$L_{2}$ &$L_{3}$ &$L_{4}$ &\\
    \midrule
    1 &0 &0 &0 &0 &182.22\\
    2 &0 &0 &0 &1 &182.18\\
    3 &0 &0 &1 &0 &181.27\\
    4 &0 &0 &1 &1 &181.53\\
    5 &0 &1 &0 &0 &182.50\\
    6 &0 &1 &0 &1 &182.12\\
    7 &0 &1 &1 &0 &181.27\\
    8 &0 &1 &1 &1 &181.38\\
    9 &1 &0 &0 &0 &183.33\\
    10 &1 &0 &0 &1 &183.32\\
    11 &1 &0 &1 &0 &182.15\\
    12 &1 &0 &1 &1 &182.14\\
    13 &1 &1 &0 &0 &183.49\\
    14 &1 &1 &0 &1 &183.40\\
    15 &1 &1 &1 &0 &182.34\\
    16 &1 &1 &1 &1 &182.42\\
    \bottomrule
  \end{tabular}
\end{table}

\subsection{cc-DRL flight control via online weighting combination}
\label{sec:Online Adaptive Control Law Performance}

The morphing quadrotor is assumed to take off with the shortest length of four arm rods. For the sake of a better flight performance, the quadrotor will expand its arm rods to the largest length. While the quadrotor must retract its arm rods to the shortest length for safely passing through two narrow channels placed at the low point of the figure-$8$ trajectory (see Fig. \ref{fig_4}). After passing through them, the quadrotor expands its arm rods to the largest length again.

\begin{figure}[!h]
  \centering
  \includegraphics[width=1\columnwidth]{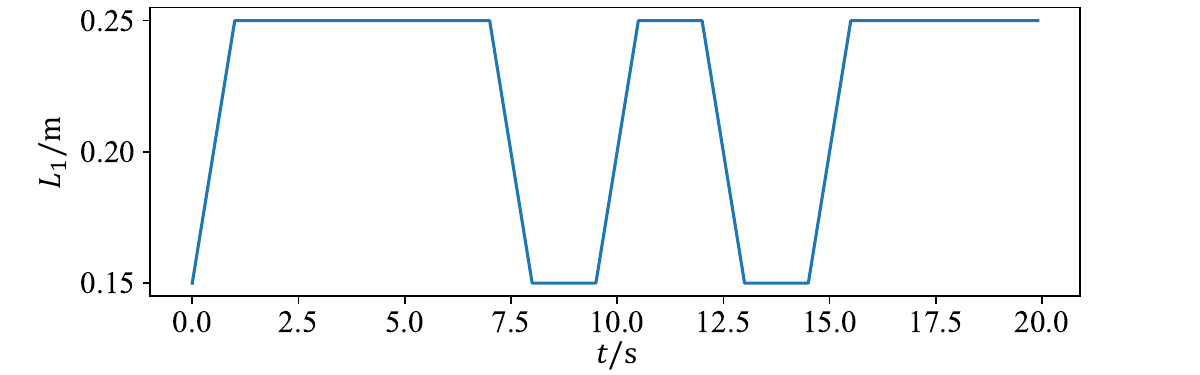}
  \includegraphics[width=1\columnwidth]{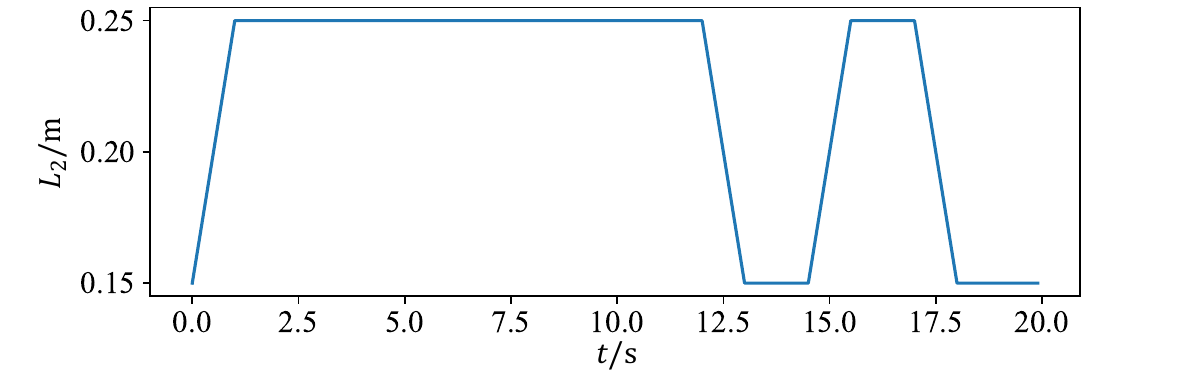}
  \includegraphics[width=1\columnwidth]{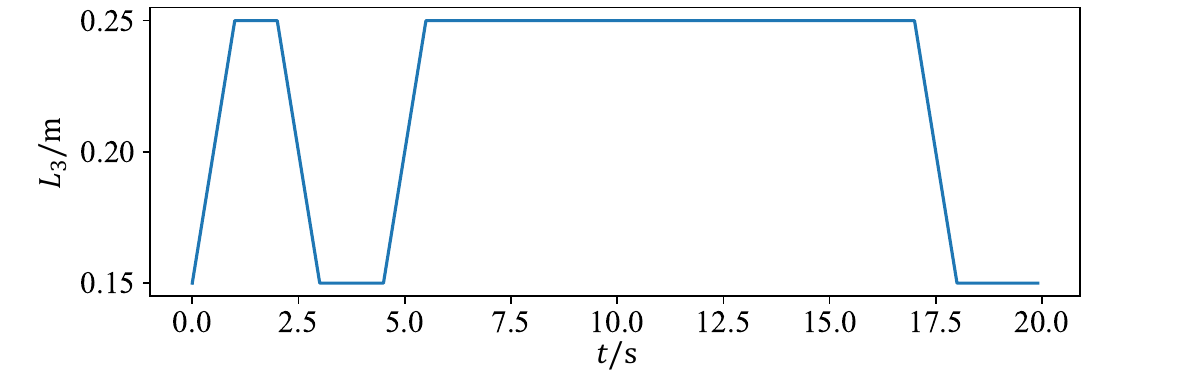}
  \includegraphics[width=1\columnwidth]{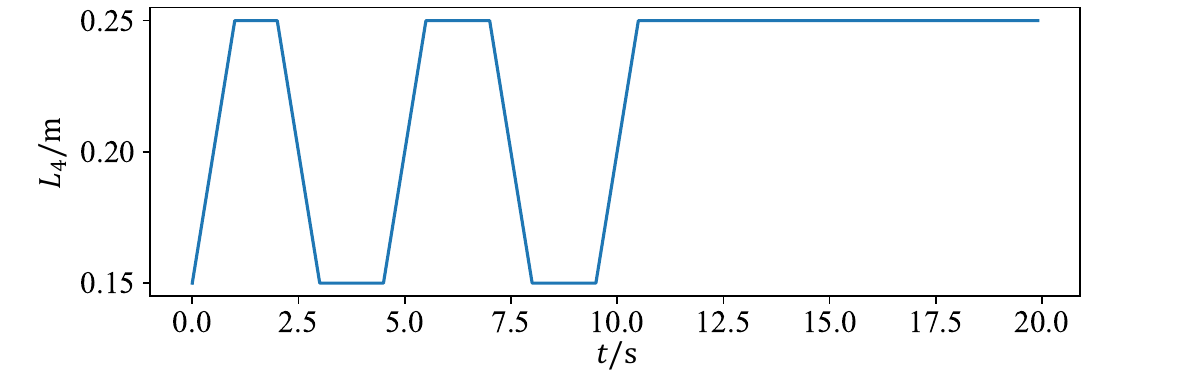}
  \caption{Trajectories of asymmetrical length changes of four arm rods}
  \label{fig_12}
\end{figure}

\begin{figure}[!t]
  \centering
  \includegraphics[width=1\columnwidth]{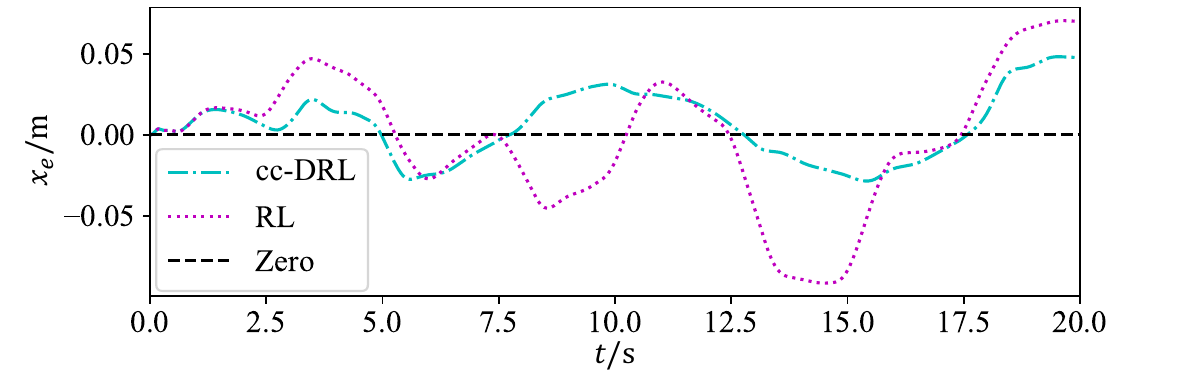}
  \includegraphics[width=1\columnwidth]{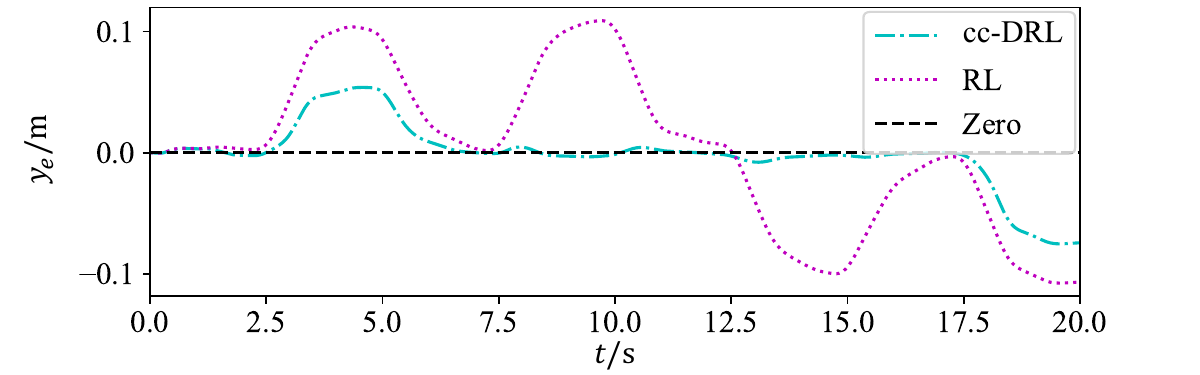}
  \includegraphics[width=1\columnwidth]{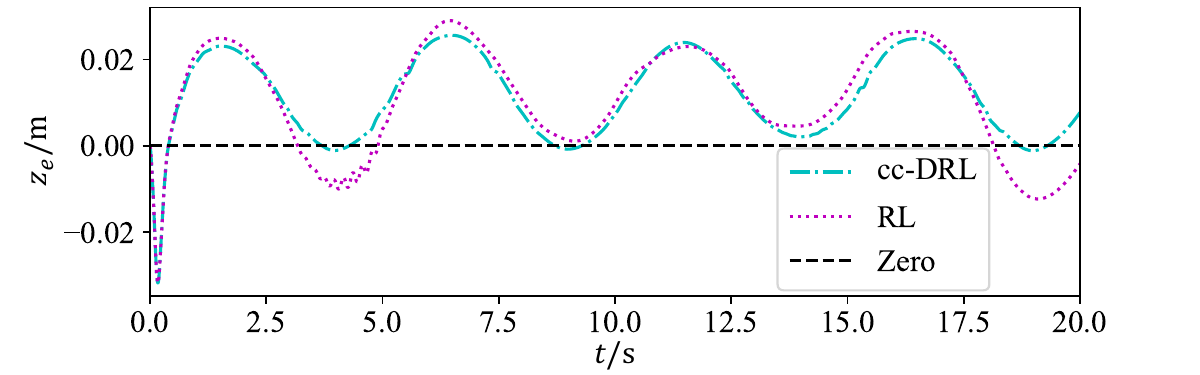}
  \caption{Position errors of figure-8 trajectory tracking for a morphing quadrotor with asymmetric length changes.}
  \label{asym_error}
\end{figure}

\begin{figure}
  \centering
  \includegraphics[width=1\columnwidth]{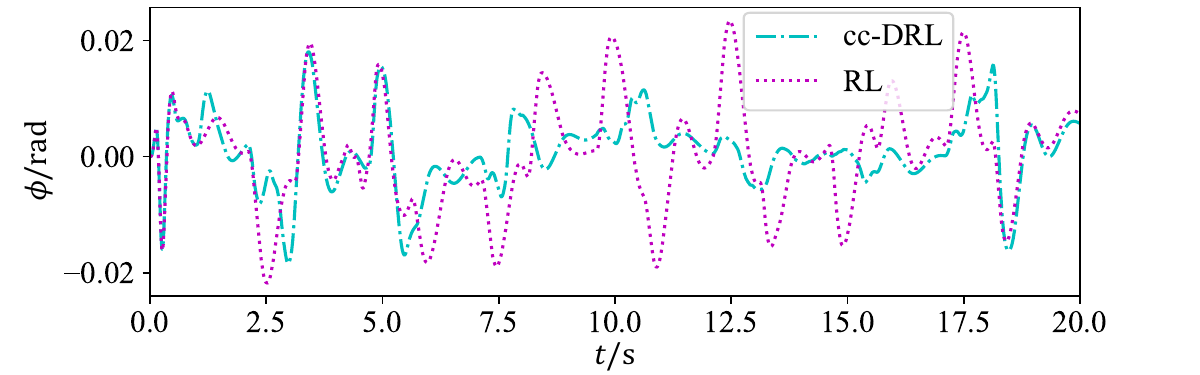}
  \includegraphics[width=1\columnwidth]{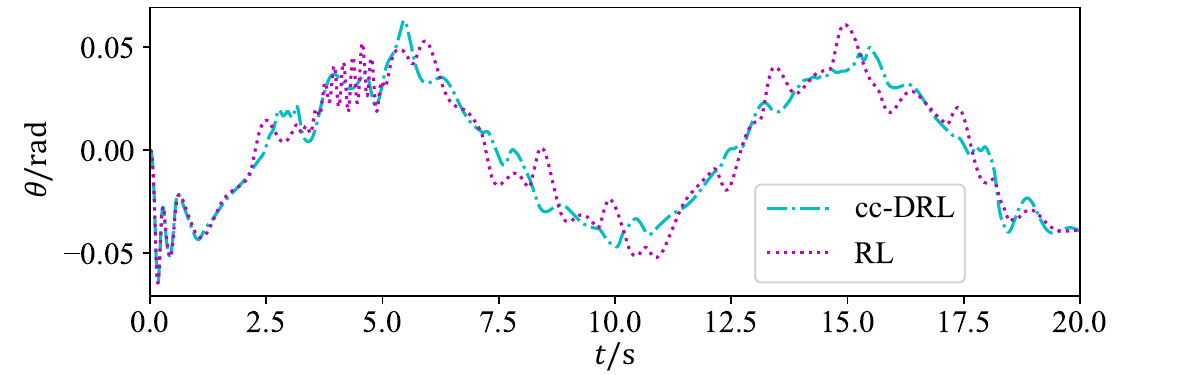}
  \includegraphics[width=1\columnwidth]{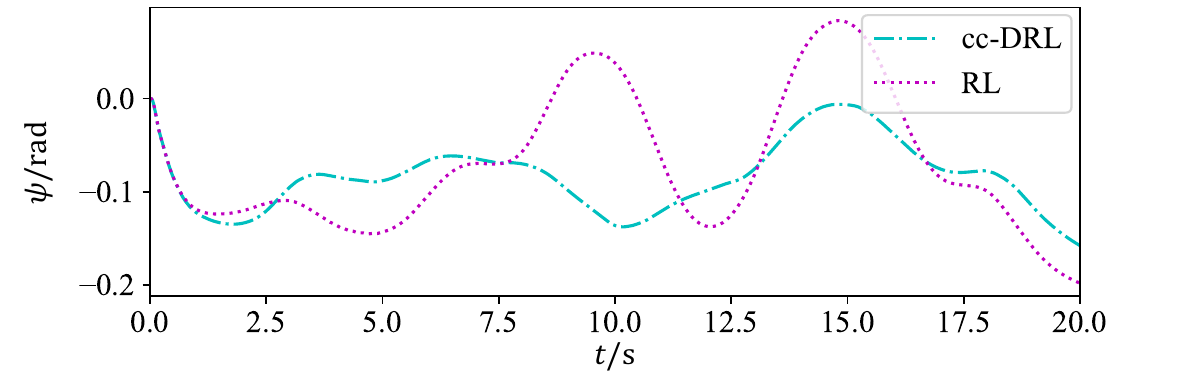}
  \caption{Attitude angle errors of figure-$8$ trajectory tracking for a morphing quadrotor with asymmetric length changes.}
  \label{asym_attitude}
\end{figure}

\begin{figure}[!t]
  \centering
  \subfloat{\includegraphics[width=0.5\columnwidth]{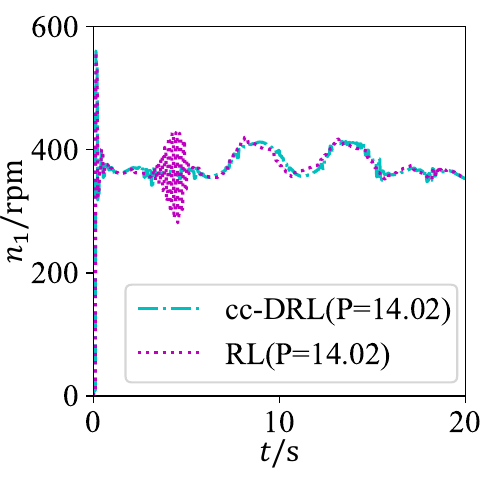}}
  \hfill
  \subfloat{\includegraphics[width=0.5\columnwidth]{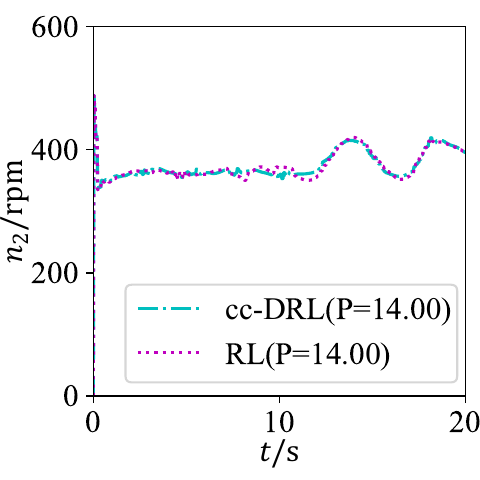}}
  \vfill
  \subfloat{\includegraphics[width=0.5\columnwidth]{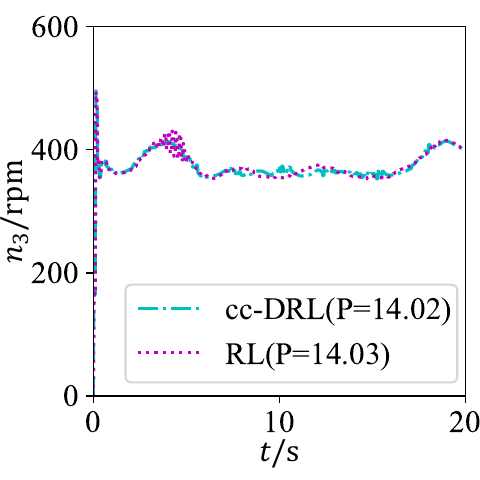}}
  \hfill
  \subfloat{\includegraphics[width=0.5\columnwidth]{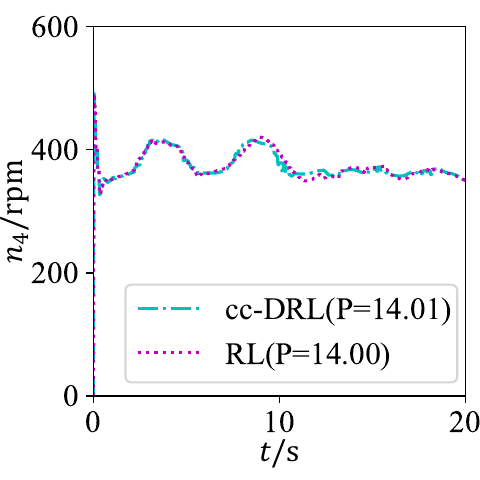}}
  \caption{Motorspeeds of figure-8 trajectory tracking for a morphing quadrotor with asymmetric length changes (The power compute by $P = \frac{1}{T}\int_{0}^{T}(\frac{n}{100})^{2}\, dt$ to measure the relative power of motors).}
  \label{asym_Motorspeed}
\end{figure}

\begin{figure}[!h]
  \centering
  \includegraphics[width=1\columnwidth]{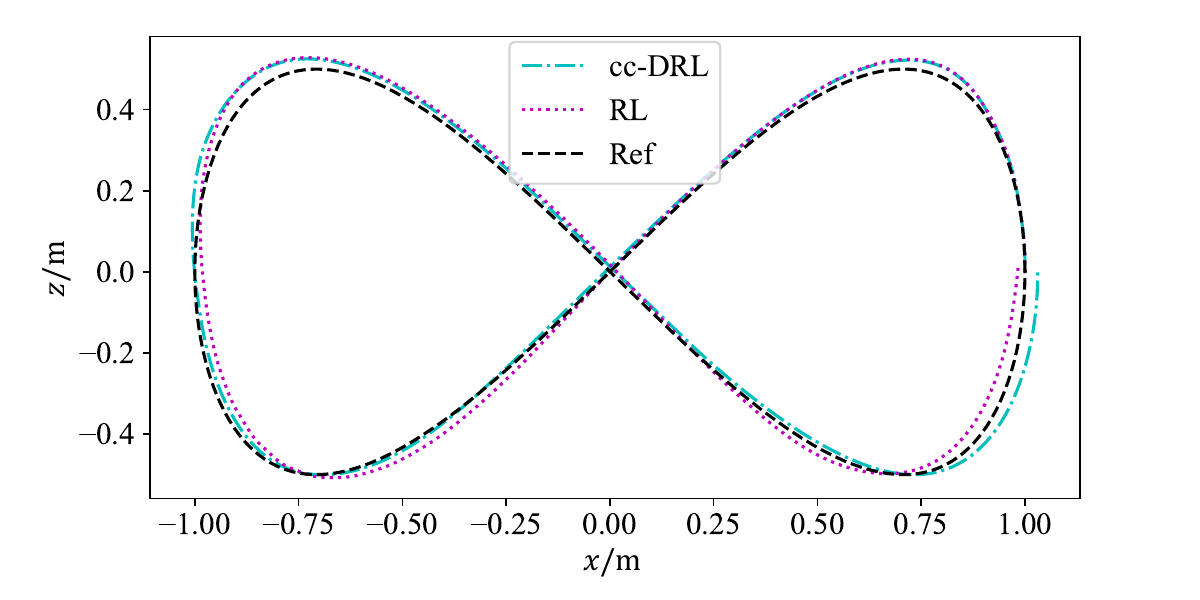}
  \caption{$2$D trajectories with narrow channels by asymmetric variation. The accumulated reward is $176.68$ for RL, and $180.69$ for CCRL, respectively.}
  \label{asym_2D}
\end{figure}

%Here, we assume that the length of four arm rods can change symmetrically as shown in Fig. \ref{fig_10}. Considering the hardware conditions, the maximum changing rate of arm length is set to be $0.1$m/s. An online overall flight control scheme is obtained by \textbf{Algorithm 3}. Time trajectories of mass center position $\textit{\textbf{x}}$ and the attitude angles $\bm{\varpi}$ of the morphing quadrotor driven by the proposed online overall flight control scheme are shown in Fig. \ref{sym_error} and Fig. \ref{sym_attitude}, respectively. Fig. \ref{sym_Motorspeed} shows the velocity of four rotors and Fig. \ref{sym_2D} gives the figure-8 flight trajectory tracking in the $xOz$ plane. The corresponding accumulated reward is 182.65 for the online overall flight control scheme.
%
%To show the advantage of the proposed overall flight control scheme, simulation results of figure-8 flight trajectory tracking are also shown in Figs. \ref{sym_error}-\ref{sym_2D}, where the morphing quadrotor is steered {by the RL scheme that is trained for the mode with four arm rod lengths of $0.15$ m.} It is clear that the proposed overall flight control scheme endows the morphing quadrotor with better flight performance.
Here, we assume that lengths of four arm rods can change asymmetrically as shown in Fig. \ref{fig_12}. Considering the hardware conditions, the maximum changing rate of arm length is set to be $0.1$m/s. A cc-DRL flight control scheme is obtained by \textbf{Algorithm 3}. Trajectories of mass center position $\textit{\textbf{x}}$ and the attitude angles $\bm{\varpi}$ for a morphing quadrotor driven by the proposed cc-DRL flight control scheme are shown in Fig. \ref{asym_error} and Fig. \ref{asym_attitude}, respectively. Fig. \ref{asym_Motorspeed} shows the velocity of four rotors and Fig. \ref{asym_2D} gives the figure-8 flight trajectory tracking in the $xOz$ plane. The corresponding accumulated reward is $180.69$ for the cc-DRL flight control scheme.

To show the advantage of the proposed cc-DRL flight control scheme, simulation results of figure-$8$ flight trajectory tracking are also shown in Figs. \ref{asym_error}-\ref{asym_2D}, where the morphing quadrotor is steered {by the RL scheme that is trained for the mode with four arm rod lengths of $0.15$ m.} The corresponding accumulated reward is $176.68$. It is clear that compared to the RL one, the proposed cc-DRL flight control scheme endows the morphing quadrotor with a better flight performance.

\section{Conclusion}
\label{sec:Conclusion}
The investigation of this study has revealed that as a model-free DRL algorithm, PPO algorithm assisted by the CC technique can effectively solve the issue of approximate optimal flight control for position and attitude of morphing quadrotors without any model knowledge of complex flight dynamics. The flight control performance of the proposed cc-DRL-based flight control algorithm is demonstrated by simulation results for an arm-rod-length-varying quadrotor. Although the proposed cc-DRL flight control algorithm is developed for a class of morphing quadrotors whose shape change is realized by the length variation of four arm rods, it is easily modified and implemented for other types of morphing quadrotors, such as tiltrotor quadrotor, multimodal quadrotor, and foldable quadrotor \cite{patnaik2021towards}.

% \section*{Acknowledgments}

\bibliographystyle{IEEEtran}
\bibliography{references}

% \section{Biography Section}

\vfill

\end{document}